\documentclass[runningheads]{llncs}

\usepackage{eccv}

\usepackage{eccvabbrv}

\usepackage{graphicx}
\usepackage{booktabs}
\usepackage{algorithm}
\usepackage{algpseudocode}
\usepackage{wrapfig}
\usepackage{xcolor}

\usepackage[accsupp]{axessibility}  

\usepackage{hyperref}
\hypersetup{hidelinks}
\usepackage{orcidlink}

\begin{document}

\title{CasDeblurGS: Cascaded 2D-to-3D Multi-View Consistency for 3D Gaussian Splatting from Two Blurry Images} 

\titlerunning{CasDeblurGS}

\author{
Haeyun Choi\inst{1}\textsuperscript{*}\textsuperscript{$\dagger$}\orcidlink{0009-0006-6399-4331}
\and
Minhyuk Jang\inst{2}\textsuperscript{*}\orcidlink{0009-0001-5549-3864}
\and
I-Gil Kim\inst{2}\orcidlink{0009-0001-4938-0038}
}

\authorrunning{H.~Choi et al.}

\institute{University of Virginia, Charlottesville, VA, USA\\
\email{phh3ps@virginia.edu}
\and
KT R\&D Center, Seoul, Republic of Korea\\
\email{\{minhyuk.jang,i-gil.kim\}@kt.com}
}

\maketitle


\begingroup
\renewcommand{\thefootnote}{*}
\footnotetext{Equal contribution.}

\renewcommand{\thefootnote}{\ensuremath{\dagger}}
\footnotetext{This work was done at KT R\&D Center.}
\endgroup

\vspace{-0.1cm}
\noindent\makebox[\linewidth][c]{%
\small
\href{https://haeyun-choi.github.io/Cascaded2D3D_page/}{%
\textcolor{magenta}{\texttt{https://haeyun-choi.github.io/Cascaded2D3D\_page/}}%
}}
\vspace{-0.8cm}

\begin{abstract}
Free-viewpoint 3D scene media is increasingly important for immersive applications, yet practical capture often suffers from severe view sparsity and motion blur.
Although neural rendering has advanced sparse-view synthesis, existing blur-aware methods typically require substantial multi-view redundancy, accurate camera poses, or costly per-scene optimization.
We address a stringent yet practical setting: reconstructing a coherent 3D scene from only two motion-blurred images with known intrinsics, without input-view poses, auxiliary sharp images, or per-scene test-time optimization.
To this end, we propose \textbf{CasDeblurGS}, a cascaded framework that progressively recovers reliable cross-view information from local 2D correspondences to global 3D guidance.
Stage~1 constructs locally reliable guidance through occlusion-aware correspondence filtering, while Stage~2 aggregates the intermediate restorations into a provisional pose-free 3D Gaussian representation whose input-view re-renders provide dense global guidance for final restoration.
The resulting views enable a more coherent 3D representation and higher-quality novel-view synthesis.
Experiments on real-world and synthetic Deblur-NeRF scenes show consistent gains over strong baselines, improving PSNR by 1.19 dB and 2.11 dB, respectively.
Progressive ablations, cross-view correspondence visualization, and camera reprojection analysis further demonstrate improvements in both rendering quality and multi-view geometric consistency.
\keywords{Gaussian Splatting \and Deblurring \and Sparse Reconstruction}
\end{abstract}

\section{Introduction}
\label{sec:introduction}
Recent advances in neural rendering, from neural radiance fields (NeRF)~\cite{mildenhall2020nerf} to 3D Gaussian Splatting (3DGS)~\cite{kerbl2023gaussiansplatting}, have substantially improved the fidelity and efficiency of novel-view synthesis.
Together with recent feed-forward reconstruction methods, which directly recover 3D representations from sparse images ~\cite{charatan2024pixelsplat,chen2024mvsplat,xu2025depthsplat, ye2024no,jiang2025anysplat,hong2024pf3plat}, these advances are making free-viewpoint 3D media increasingly practical for applications such as XR, immersive telepresence, and digital content creation~\cite{bao20253d,joshi2025unconstrained,gao2024cat3d, khan2025autosplat,zhang2025egogaussian}.

Despite this progress, practical capture remains challenging.
Dense multi-view acquisition may be infeasible because of limited capture time, sensing conditions, or platform stability, forcing reconstruction to rely on only a few observations.
When these sparse inputs are further degraded by motion blur from handheld imaging, platform vibration, or low-light exposure, image details and cross-view correspondences are simultaneously corrupted, making reliable scene recovery substantially more difficult.

\begin{figure}[t]
    \centering
    \includegraphics[width=\linewidth]{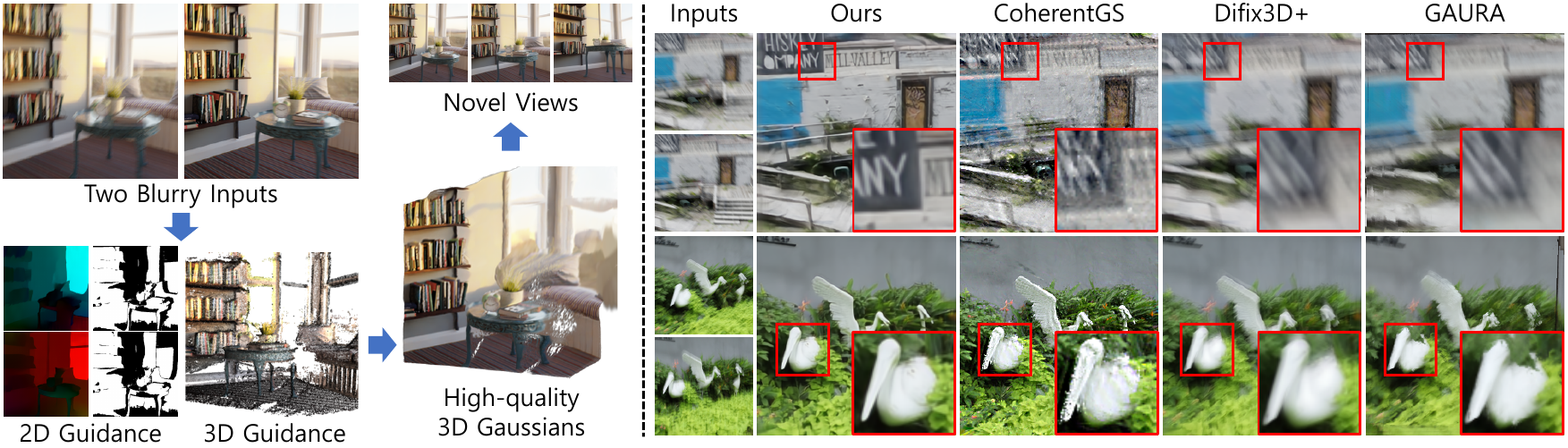}
    \caption{
    (\emph{Left}) Given two blurry images, CasDeblurGS progressively combines local 2D guidance and global 3D guidance for coherent 3D Gaussian reconstruction.
    (\emph{Right}) Compared to CoherentGS~\cite{xu2025breaking}, Difix3D+~\cite{wu2025difix3d+}, and GAURA~\cite{gupta2024gaura}, ours produces sharper details and more coherent novel views in the challenging two-view blurry setting.
    }
    \label{fig:teaser}
\end{figure}

Recent studies have tackled this challenge by incorporating blur-aware modeling into neural rendering or jointly optimizing image restoration and scene representation~\cite{ma2022deblurnerf,lee2024deblurring,peng2024bags,zhao2024bad,choi2025exploiting,zhao2025bsgs}.
However, these approaches remain difficult to deploy in practice.
First, they typically rely on per-scene test-time optimization and substantial multi-view redundancy to stabilize the joint estimation of blur and scene structure, often requiring 20--40 input images.
Such assumptions are difficult to satisfy when only a few blurry views are available and rapid reconstruction is desired.
Second, they depend on reliable geometric initialization under blur, typically in the form of accurate camera poses or coarse scene geometry.
Such initialization is already difficult to recover from blurry frames using standard structure-from-motion pipelines~\cite{schonberger2016structure}.
When both view redundancy and geometric initialization are weak, the problem becomes ill-posed, often leading to unstable and spurious geometry.

These limitations become even more severe in sparse-view settings, where blur further weakens the already limited cross-view constraints.
Consequently, even methods designed for sparse-view blurry reconstruction remain fragile in the extreme two-view regime~\cite{lee2025sparse,xu2025breaking}.
Under such limited observations, the joint estimation of blur and scene structure becomes severely ill-conditioned, often yielding blurry renderings, noisy geometry, and floating artifacts.

In this work, we consider a stringent yet practical setting: reconstructing a coherent 3D scene from only two motion-blurred images with known camera intrinsics, without input-view camera poses, auxiliary
sharp images, or per-scene test-time optimization.
The central difficulty is that severe blur corrupts the already limited cross-view evidence required for both local correspondence estimation and global 3D aggregation.
Directly transferring unreliable 2D correspondences can introduce misaligned structures, while aggregating inconsistent observations in 3D can propagate these errors into the reconstructed scene.

Our key idea is to recover reliable cross-view information \emph{progressively}, from local 2D correspondences to global 3D guidance.
We propose \textbf{CasDeblurGS}, a cascaded 2D-to-3D multi-view consistency framework.
In Stage~1, a frozen stabilizer first produces alignment-friendly observations, after which our occlusion-aware cross-view guidance module (OCGM) filters unreliable optical-flow correspondences through forward--backward consistency and constructs masked reference warps for intermediate restoration.
In Stage~2, these restorations are aggregated by a frozen pose-free 3DGS backbone into a provisional 3D representation whose input-view re-renders provide dense global guidance for final restoration.
The final restored views are then passed through the same frozen backbone to construct the output 3D representation for novel-view synthesis.
At inference time, CasDeblurGS requires only the two blurry images and their intrinsics, with no camera poses, depth maps, external generative priors, or scene-specific optimization.

This progressive design combines complementary strengths of 2D and 3D guidance.
Local warping selectively transfers fine cross-view evidence where correspondences are reliable, whereas the provisional 3D representation aggregates information from both views to provide dense scene-level
feedback. 
Consequently, the cascade improves both the quality of the restored observations and their consistency for downstream 3D reconstruction.

Experiments on synthetic and real-world Deblur-NeRF scenes demonstrate consistent gains over strong blur-aware and restoration-based baselines.
CasDeblurGS improves PSNR over the strongest competing results by 1.19~dB on real-world scenes and 2.11~dB on synthetic scenes. 
Progressive ablations, cross-view correspondence visualization, and camera reprojection analysis further demonstrate improvements in both novel-view synthesis quality and multi-view geometric consistency.

Our contributions are summarized as follows:
\begin{itemize}
    \item We introduce CasDeblurGS, a pose-free feed-forward framework for 3D reconstruction from only two motion-blurred images with known intrinsics, without auxiliary sharp images or per-scene test-time
    optimization.
    \item We propose a cascaded 2D-to-3D guidance strategy that establishes locally reliable cross-view correspondences through OCGM and then exploits pose-free 3D re-rendering to provide dense global
    guidance for final restoration.
    \item We demonstrate consistent improvements on real-world and synthetic Deblur-NeRF scenes, supported by progressive ablations, cross-view correspondence visualization, and camera reprojection analysis.
\end{itemize}

\section{Related Work}
\label{sec:related_work}
\subsection{Radiance Fields from Blurry Images}
Recent advances in NeRF and 3D Gaussian Splatting (3DGS) have spurred research on modeling camera motion blur for sharp 3D scene reconstruction. 
A common strategy is to simulate the blur formation process by modeling camera motion during exposure as a continuous trajectory. 
Existing methods parameterize this trajectory using SE(3) interpolation for linear motion~\cite{wang2023bad, zhao2024bad, wu2024deblur4dgs}, Bézier curves for complex nonlinear paths~\cite{lee2023exblurf}, or Neural ODEs for flexible temporal transformations~\cite{lee2024crim, lee2025comogaussian}. 
However, such trajectory optimization typically requires substantial multi-view redundancy to jointly stabilize blur and geometry estimation. 
In sparse-view settings, insufficient multi-view constraints make this estimation unreliable; in the extreme two-view blurry regime, standard SfM initialization becomes infeasible and cross-view correspondences are severely weakened, leading to unstable geometry estimation.

Alternatively, some methods address sparse blurry inputs using 2D structural or semantic priors, or external generative models. 
For instance, HQGS~\cite{lin2025hqgs} leverages 2D edges and semantic cues, S2Gaussian~\cite{wan2025s2gaussian} resolves feature-space cross-view inconsistencies during 2D super-resolution, and CoherentGS~\cite{xu2025breaking} augments virtual views with deblurring networks and diffusion priors. However, these methods rely heavily on local 2D cues or external priors. In extreme two-view settings with limited geometric constraints, they struggle to ensure global 3D coherence, often producing structural distortions and cross-view inconsistencies in the final renderings.

\subsection{Feed-forward 3D Gaussian Splatting}
While traditional 3DGS relies on per-scene optimization, recent generalizable feed-forward methods~\cite{zhang2025advances} directly predict 3D Gaussians from sparse views. 
Building on earlier generalizable NeRFs~\cite{yu2021pixelnerf, wang2021ibrnet, chen2021mvsnerf}, 3DGS-based models now dominate this direction, using epipolar geometry, cost volumes, depth-aware matching, or hybrid volume-pixel representations for cross-view aggregation~\cite{charatan2024pixelsplat, chen2024mvsplat, zhang2025pansplat, xu2025depthsplat, wei2025omni, tang2024hisplat}. 
Pose-free variants further remove explicit pose dependence through coarse-to-fine alignment or canonical-space prediction~\cite{jiang2025anysplat, ye2024no, hong2024pf3plat}. 
However, these methods largely assume sharp inputs and remain vulnerable to severe motion blur.

GAURA~\cite{gupta2024gaura} jointly addresses feed-forward reconstruction and deblurring, but it assumes known camera poses, uses kernel-based synthetic blur, and remains unverified in the extreme two-view regime. 
More fundamentally, feed-forward aggregation depends on reliable cross-view correspondence cues, such as photometric consistency or epipolar features, which degrade severely under strong blur. 
This problem is amplified in two-view settings, where no additional observations are available to compensate for the lost constraints.

\subsection{Multi-view Image Deblurring}
Multi-view deblurring exploits geometric cues such as disparity and depth for restoration~\cite{zhou2019davanet,yan2020disparity,pan2017simultaneous}, but its reliance on correspondence estimation limits its effectiveness in extreme two-view blurry settings where high-frequency details are severely degraded and reliable camera poses and depth are unavailable.

Fundamentally, improving 2D image quality alone does not guarantee 3D consistency for novel view synthesis~\cite{ma2022deblurnerf,wang2023bad,lee2023dpnerf,lee2023exblurf}. 
Recent diffusion-based approaches aim at 3D-consistent restoration~\cite{wu2025difix3d+,mao2025sir,luo20253denhancer}, yet remain vulnerable when severe blur collapses geometric matching cues.
Motivated by these limitations, we propose a cascaded design that progresses from flow-based local 2D guidance to global 3D constraints via pose-free 3DGS re-rendering, enabling stable reconstruction without external generative priors at inference time.

\section{Preliminaries}
\label{sec:preliminaries}
\subsection{3D Gaussian Splatting}
3D Gaussian Splatting (3DGS) represents a scene using a set of anisotropic 3D Gaussians:
\begin{equation}
\mathcal{G}=\{g_m\}_{m=1}^{M}, \qquad
g_m=(\boldsymbol{\mu}_m,\mathbf{q}_m,\mathbf{s}_m,\alpha_m,\mathbf{c}_m),
\end{equation}
where $\boldsymbol{\mu}_m \in \mathbb{R}^3$ denotes the Gaussian center, $\mathbf{q}_m$ the rotation, $\mathbf{s}_m$ the scale, $\alpha_m$ the opacity, and $\mathbf{c}_m$ the appearance parameters, e.g., spherical harmonics coefficients~\cite{kerbl2023gaussiansplatting,ye2024no}. 
Given a target viewpoint $v_t$, the differentiable 3DGS rasterizer renders an image as:
\begin{equation}
\hat{I}_t=\mathrm{Render}(\mathcal{G}, v_t).
\label{eq:render}
\end{equation}
This formulation supports efficient differentiable rendering and serves as the scene representation used throughout our pipeline.

\subsection{Pose-free Feed-forward 3DGS}
Pose-free feed-forward 3DGS reconstructs a scene directly from sparse images and their camera intrinsics without requiring input-view camera extrinsics.
In particular, NoPoSplat~\cite{ye2024no} learns a mapping:
\begin{equation}
\mathcal{G}=h_{\eta}\big(\{(I_i,K_i)\}_{i=1}^{N}\big),
\end{equation}
where $I_i$ and $K_i$ denote the $i$-th input image and its camera intrinsics, respectively. 
NoPoSplat uses the first input view to establish a canonical coordinate frame and directly predicts the Gaussians associated with all input
views in this shared space.
This enables direct multi-view fusion without externally supplied input-view camera poses.
The camera intrinsics are incorporated into the input representation to account for camera geometry and alleviate scene-scale ambiguity.
The resulting Gaussian representation can then be rendered from a target viewpoint expressed in the same canonical frame using
\cref{eq:render}.

In this work, we instantiate the backbone in the two-view setting ($N=2$) and keep $h_{\eta}$ frozen throughout the pipeline.
As detailed in \cref{sec:method}, its canonical-space reconstruction and re-rendering capabilities are used for global 3D guidance and
final novel-view synthesis, without scene-specific test-time optimization.

\section{Method}
\label{sec:method}

\begin{figure}[t] 
    \centering
    \includegraphics[width=\textwidth]{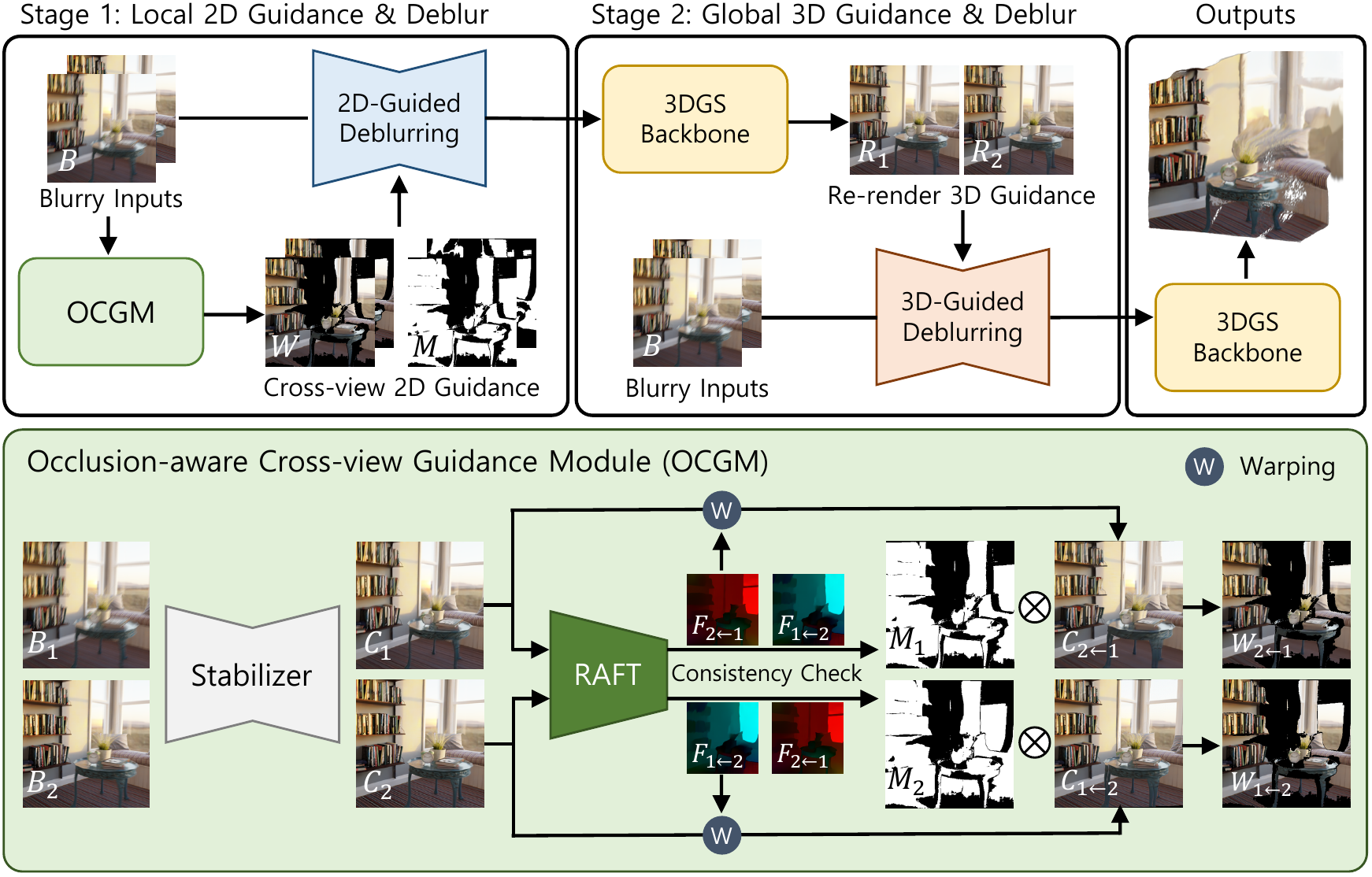}
    \caption{
    Overview of CasDeblurGS. 
    (Top) The cascade constructs local 2D guidance, then uses a frozen pose-free 3DGS backbone to provide global 3D guidance for final restoration. 
    Final restorations feed the same backbone for novel-view synthesis.
    (Bottom) OCGM stabilizes inputs, estimates bidirectional RAFT flow, derives forward--backward consistency masks, and produces masked cross-view warps for Stage~1.
    }
    \label{fig:framework}
\end{figure}

Given two motion-blurred images and their known camera intrinsics, $\{(B_i,K_i)\}_{i=1}^{2}$, CasDeblurGS progressively recovers
reliable cross-view guidance to restore the input views and construct a pose-free 3D Gaussian representation for novel-view synthesis.
No ground-truth or pre-estimated input-view camera poses are provided to the framework.
As illustrated in \cref{fig:framework}, the pipeline comprises two cascaded restoration stages followed by final pose-free 3D reconstruction.

Stage~1 establishes locally reliable correspondence-based guidance and uses it to produce sharper intermediate restorations with improved cross-view consistency.
Stage~2 aggregates these intermediate views into a provisional 3D representation and re-renders it at the input viewpoints, providing
dense global guidance for final restoration.
The final restored views are then processed by the same frozen 3DGS backbone to construct the output representation for novel-view synthesis.

The stabilizer $\mathcal{S}_{\phi}$~\cite{chen2022simple}, the RAFT-based optical-flow estimator~\cite{teed2020raft}, and the pose-free 3DGS backbone $h_{\eta}$~\cite{ye2024no} remain frozen throughout the pipeline.
The 2D- and 3D-guided restoration networks, $\mathcal{D}^{2D}_{\theta}$ and $\mathcal{D}^{3D}_{\psi}$, are trained offline in a stage-wise manner.
All modules are fixed at inference, and CasDeblurGS therefore requires no scene-specific test-time optimization.

The key design of CasDeblurGS is the progressive construction of reliable cross-view guidance.
Stage~1 suppresses unreliable local correspondences through occlusion-aware masking, whereas Stage~2 lifts the intermediate restorations into a shared 3D representation and projects the aggregated scene information back to the image plane.
\Cref{alg:cascade_2d3d} summarizes the complete inference pipeline.
Training and implementation details are provided in \cref{subsec:experimental_setup} and the supplementary material.

\subsection{Stage~1: Local 2D Guidance and Deblurring}
\label{sec:stage1}

Stage~1 constructs locally reliable correspondence-based guidance from the two input views.
It transfers cross-view information only where the estimated correspondence is sufficiently trustworthy.
The stage consists of alignment preconditioning, occlusion-aware cross-view guidance generation, and 2D-guided restoration.

\paragraph{\normalfont\bfseries Alignment preconditioning with a Stabilizer.}
Directly estimating optical flow between severely motion-blurred inputs is unreliable because blur suppresses and distorts structures
shared across views.
We therefore first process each input using a frozen pretrained stabilizer:
\begin{equation}
C_1 = \mathcal{S}_{\phi}(B_1),
\qquad
C_2 = \mathcal{S}_{\phi}(B_2).
\end{equation}
The stabilized views $(C_1,C_2)$ are not treated as final restorations.
Instead, they serve as alignment-friendly observations from which more reliable cross-view correspondences can be estimated.

\paragraph{\normalfont\bfseries Occlusion-aware cross-view guidance.}
From the stabilized views, we estimate bidirectional optical flow using a frozen RAFT model~\cite{teed2020raft}:
\begin{equation}
F_{1\leftarrow2} = \mathrm{RAFT}(C_2,C_1),
\;
F_{2\leftarrow1} = \mathrm{RAFT}(C_1,C_2).
\end{equation}
Here, $F_{b\leftarrow r}$ denotes a flow field defined on the base-view coordinates of $b$, used to map content from the reference view $r$ into view $b$.
Not every estimated correspondence provides reliable restoration guidance.
Occlusions, motion boundaries, out-of-bounds projections, and blur-induced flow errors can introduce misaligned or duplicated
structures.
We therefore validate each correspondence using forward--backward consistency.
For a pixel location $x$ in the base view, its mapped coordinate in the reference view is
\begin{equation}
x' = x + F_{b\leftarrow r}(x).
\end{equation}
The corresponding forward--backward residual is
\begin{equation}
e(x)
=
\left\|
F_{b\leftarrow r}(x)
+
F_{r\leftarrow b}(x')
\right\|_2.
\end{equation}
To avoid over-rejecting valid correspondences under large displacements, we use the motion-adaptive threshold

\begin{equation}
T(x)
=
\tau
+
\alpha
\left(
\left\|F_{b\leftarrow r}(x)\right\|_2
+
\left\|F_{r\leftarrow b}(x')\right\|_2
\right),
\end{equation}
where $\tau$ provides a base tolerance for small estimation and resampling errors, and $\alpha$ adjusts the tolerance according to
the motion magnitude.
The validity mask is defined as
\begin{equation}
M_{b\leftarrow r}(x)
=
\mathbb{1}[\mathrm{in\text{-}bounds}(x')]
\cdot
\mathbb{1}[e(x) \leq T(x)].
\end{equation}
Only correspondences satisfying both the in-bounds and forward--backward consistency conditions are retained.
We then construct the masked cross-view warp
\begin{equation}
W_{b\leftarrow r}
=
M_{b\leftarrow r}
\odot
\mathcal{W}\bigl(C_r,F_{b\leftarrow r}\bigr),
\end{equation}
where $\mathcal{W}$ denotes backward warping.
We refer to this combination of bidirectional flow estimation, forward--backward validation, and masked warping as the \emph{occlusion-aware cross-view guidance module} (OCGM).
Applying OCGM in both directions produces the guidance--mask pairs $(W_{1\leftarrow2},M_{1\leftarrow2})$ and
$(W_{2\leftarrow1},M_{2\leftarrow1})$.

\paragraph{\normalfont\bfseries 2D-guided deblurring.}
For each view, we concatenate the original blurry input, the corresponding masked cross-view warp, and its validity mask:
\begin{equation}
\begin{aligned}
\tilde{S}_1
&=
\mathcal{D}^{2D}_{\theta}
\bigl(
\mathrm{concat}
(B_1,W_{1\leftarrow2},M_{1\leftarrow2})
\bigr), \\
\tilde{S}_2
&=
\mathcal{D}^{2D}_{\theta}
\bigl(
\mathrm{concat}
(B_2,W_{2\leftarrow1},M_{2\leftarrow1})
\bigr).
\end{aligned}
\end{equation}
Using the original blurry image as the base preserves image evidence that may be altered by the stabilizer, while the masked warp transfers
complementary information from the other view.
The explicit validity mask additionally indicates where the cross-view guidance can be trusted.
The resulting intermediate restorations $(\tilde{S}_1,\tilde{S}_2)$ provide sharper observations with improved cross-view consistency for constructing the provisional 3D representation in Stage~2.

\subsection{Stage~2: Global 3D Guidance and Deblurring}
\label{sec:stage2}

Stage~2 aggregates the intermediate restorations into a provisional 3D representation and projects the resulting scene-level information
back to the image plane for final restoration.
While Stage~1 transfers locally reliable information through correspondence-based masked warping, Stage~2 provides dense global
guidance induced by a shared 3D representation.

\paragraph{\normalfont\bfseries Provisional 3D representation and re-render guidance.}
We feed the intermediate restorations $(\tilde{S}_1,\tilde{S}_2)$ and their camera intrinsics into the frozen pose-free 3DGS backbone:
\begin{equation}
\mathcal{G} = h_{\eta}\big((\tilde{S}_1,K_1), (\tilde{S}_2,K_2)\big).
\end{equation}
We then re-render the resulting 3D Gaussian representation at the two input viewpoints:
\begin{equation}
R_1 = \mathrm{Render}(\mathcal{G},v_1), \qquad
R_2 = \mathrm{Render}(\mathcal{G},v_2),
\end{equation}
where $v_1$ denotes the first input viewpoint in the canonical camera frame established by the pose-free backbone, and $v_2$ is determined
by the relative camera geometry inferred for the second view.
Both viewpoints are internal to the backbone's canonical-space formulation; no ground-truth or pre-estimated input-view poses are
provided to CasDeblurGS.
Because $\mathcal{G}$ is jointly constructed from both intermediate restorations, the re-rendered images $(R_1,R_2)$ encode scene
information aggregated across the two views and provide dense 3D guidance over the full image plane.

\paragraph{\normalfont\bfseries 3D-guided deblurring.}
For each view, we combine the original blurry input with its corresponding re-render guidance:
\begin{equation}
\begin{aligned}
\hat{S}_1
&=
\mathcal{D}^{3D}_{\psi}
\bigl(
\mathrm{concat}(B_1,R_1)
\bigr), \\
\hat{S}_2
&=
\mathcal{D}^{3D}_{\psi}
\bigl(
\mathrm{concat}(B_2,R_2)
\bigr).
\end{aligned}
\end{equation}
The original blurry input preserves view-specific image evidence, while the re-rendered view supplies complementary scene-level
structure aggregated through the shared 3D representation.
Unlike the spatially selective correspondence guidance used in Stage~1, the re-rendered images provide dense guidance across the
full image plane.
Stage~2 thereby produces the final restored views $(\hat{S}_1,\hat{S}_2)$ using information jointly derived from both intermediate restorations.

\subsection{Final 3D Representation for NVS}
\label{sec:final3d}
We apply the same frozen pose-free 3DGS backbone once more to the final restored views:
\begin{equation}
\mathcal{G}^{\star}
=
h_{\eta}\big(
(\hat{S}_1,K_1),
(\hat{S}_2,K_2)
\big).
\end{equation}
The provisional representation $\mathcal{G}$ is constructed from the intermediate restorations solely to generate the re-render guidance
used in Stage~2.
In contrast, $\mathcal{G}^{\star}$ is constructed from the final restored views and serves as the output 3D representation for novel-view synthesis.

Given $\mathcal{G}^{\star}$, we render novel viewpoints using the differentiable 3DGS rasterizer.
CasDeblurGS performs no per-scene or test-time optimization of $\mathcal{G}^{\star}$.
Instead, the proposed cascade supplies the frozen pose-free backbone with sharper and more mutually consistent input observations,
enabling it to construct a more coherent final 3D representation.

\begin{algorithm}[t]
\caption{Inference Pipeline of CasDeblurGS}
\label{alg:cascade_2d3d}
\small
\begin{algorithmic}[1]

\Require Two blurry images and intrinsics
$\{(B_i,K_i)\}_{i=1}^{2}$
\Ensure Restored images $(\hat{S}_1,\hat{S}_2)$ and
3D representation $\mathcal{G}^{\star}$

\Statex \textbf{Note.} All modules are frozen at inference.

\Statex \textbf{Stage~1: Local 2D Guidance and Deblurring}
\State $C_1 \leftarrow \mathcal{S}_{\phi}(B_1)$ \Comment{stabilization}
\State $C_2 \leftarrow \mathcal{S}_{\phi}(B_2)$
\State $F_{1\leftarrow 2} \leftarrow \mathrm{RAFT}(C_2, C_1)$ \Comment{optical flow}
\State $F_{2\leftarrow 1} \leftarrow \mathrm{RAFT}(C_1, C_2)$
\State $M_{1\leftarrow 2} \leftarrow \mathrm{FBMask}(F_{1\leftarrow 2}, F_{2\leftarrow 1})$ \Comment{validity mask}
\State $M_{2\leftarrow 1} \leftarrow \mathrm{FBMask}(F_{2\leftarrow 1}, F_{1\leftarrow 2})$
\State $W_{1\leftarrow 2} \leftarrow M_{1\leftarrow 2} \odot \mathcal{W}(C_2, F_{1\leftarrow 2})$ \Comment{masked warp}
\State $W_{2\leftarrow 1} \leftarrow M_{2\leftarrow 1} \odot \mathcal{W}(C_1, F_{2\leftarrow 1})$
\State $\tilde{S}_1 \leftarrow \mathcal{D}^{2D}_{\theta}(\mathrm{concat}(B_1, W_{1\leftarrow 2}, M_{1\leftarrow 2}))$ \Comment{2D guidance}
\State $\tilde{S}_2 \leftarrow \mathcal{D}^{2D}_{\theta}(\mathrm{concat}(B_2, W_{2\leftarrow 1}, M_{2\leftarrow 1}))$

\Statex \textbf{Stage~2: Global 3D Guidance and Deblurring}
\State $\mathcal{G} \leftarrow h_\eta\big((\tilde{S}_1, K_1), (\tilde{S}_2, K_2)\big)$ \Comment{pose-free 3DGS backbone}
\State $R_1 \leftarrow \mathrm{Render}(\mathcal{G}, v_1)$ \Comment{re-render}
\State $R_2 \leftarrow \mathrm{Render}(\mathcal{G}, v_2)$
\State $\hat{S}_1 \leftarrow \mathcal{D}^{3D}_{\psi}(\mathrm{concat}(B_1, R_1))$ \Comment{3D guidance}
\State $\hat{S}_2 \leftarrow \mathcal{D}^{3D}_{\psi}(\mathrm{concat}(B_2, R_2))$

\Statex \textbf{Final 3D Representation for Novel View Synthesis}
\State $\mathcal{G}^\star \leftarrow h_\eta\big((\hat{S}_1, K_1), (\hat{S}_2, K_2)\big)$
\State \Return $(\hat{S}_1, \hat{S}_2, \mathcal{G}^\star)$
\end{algorithmic}
\end{algorithm}

\section{Experiments}
\label{sec:experiments}

\subsection{Experimental Setup}
\label{subsec:experimental_setup}
\paragraph{\normalfont\bfseries Datasets and two-view evaluation protocol.}
We train our restoration networks on the synthetic camera-motion-blur dataset introduced in DeepDeblurRF~\cite{choi2025exploiting} and evaluate on the camera-motion-blur subset of Deblur-NeRF~\cite{ma2022deblurnerf}, which comprises five synthetic scenes and ten real-world scenes.
For a fair comparison, we evaluate all methods on 85 fixed tuples across 15 scenes: 25 synthetic and 60 real-world.
Each tuple contains two blurry inputs and one held-out target view for novel-view synthesis.
The complete input--target mappings are provided in the supplementary material.
All images are resized with preserved aspect ratio and center-cropped to $256\times256$ to match the pose-free 3DGS backbone~\cite{ye2024no}.

\paragraph{\normalfont\bfseries Compared methods.}
We compare CasDeblurGS with representative sparse-view 3DGS methods.
SE-GS~\cite{zhao2025self} performs per-scene optimization for few-shot novel-view synthesis, while GAURA~\cite{gupta2024gaura} is a generalizable feed-forward method for sparse blurry inputs.
CoherentGS~\cite{xu2025breaking} targets sparse motion blur using video diffusion priors and alternating per-scene optimization.

To test whether stronger 2D restoration before reconstruction is sufficient, we construct two deblurring-then-reconstruction baselines.
Specifically, we pair the same frozen pose-free 3DGS backbone used in CasDeblurGS with DAVANet~\cite{zhou2019davanet}, a stereo deblurring network, and Difix3D+~\cite{wu2025difix3d+}, a diffusion-based 3D enhancement method.
For these controls, the ``Pose-Free'' and ``Generalizable'' refer to the integrated pipelines rather than to the restoration models alone.
All methods use the same two-view protocol and input--target mappings (see Supplementary Material), with official implementations adapted where necessary.

\paragraph{\normalfont\bfseries Implementation details.}
We use a pretrained NAFNet~\cite{chen2022simple} as the frozen stabilizer $\mathcal{S}_{\phi}$ and RAFT-Large~\cite{teed2020raft} as
the frozen optical-flow estimator.
For $h_{\eta}$, we use the NoPoSplat checkpoint~\cite{ye2024no} trained on RealEstate10K with two $256\times256$ inputs.
Both restoration networks, $\mathcal{D}^{2D}_{\theta}$ and $\mathcal{D}^{3D}_{\psi}$, follow the NAFNet architecture with a base width of 64.

Stage~1 takes a 7-channel concatenation of the blurry base view, masked cross-view warp, and validity mask, whereas Stage~2 takes a 6-channel concatenation of the blurry base view and its 3D re-render guidance.
We train the two restoration networks stage-wise.
The Stage~1 network is trained first, after which it is fixed while training the Stage~2 network.

Each stage is optimized for 400K iterations using AdamW with an initial learning rate of $10^{-3}$, weight decay of $10^{-3}$, and
$\beta_1=\beta_2=0.9$.
We use cosine learning-rate decay with a minimum learning rate of $10^{-7}$ and a batch size of 32 per GPU on two NVIDIA A100 GPUs.
The training objective combines an $\ell_1$ reconstruction loss and a VGG-based perceptual loss~\cite{simonyan2014very,wang2021realesrgan}.
For OCGM, we set the forward--backward consistency parameters to $\tau=1.0$ and $\alpha=0.01$ in all experiments.
Sensitivity analysis for these parameters, runtime comparisons, and additional implementation details are provided in the supplementary material.

\begin{table}[!t]
\caption{
Novel-view synthesis results on real-world and synthetic Deblur-NeRF scenes.
DAVANet and Difix3D+ use the same frozen pose-free 3DGS backbone as ours.
Best and second-best are highlighted in \textbf{bold} and \underline{underlined}, respectively.
}
\label{tab:main_results}
\centering
\setlength{\tabcolsep}{3pt}
\renewcommand{\arraystretch}{1.05}
\resizebox{\textwidth}{!}{%
\begin{tabular}{@{}lcccccccc@{}}
\toprule
& & &
\multicolumn{3}{c}{Real-world} &
\multicolumn{3}{c}{Synthetic} \\
\cmidrule(lr){4-6}
\cmidrule(lr){7-9}
Method
& Pose-Free
& Generalizable
& PSNR $\uparrow$
& SSIM $\uparrow$
& LPIPS $\downarrow$
& PSNR $\uparrow$
& SSIM $\uparrow$
& LPIPS $\downarrow$ \\
\midrule

SE-GS~\cite{zhao2025self}
& $\times$ & $\times$
& 15.90 & 0.398 & 0.443
& 18.76 & 0.560 & 0.338 \\

GAURA~\cite{gupta2024gaura}
& $\times$ & \checkmark
& 17.67 & 0.508 & 0.444
& 17.91 & 0.524 & 0.414 \\

CoherentGS~\cite{xu2025breaking}
& $\times$ & $\times$
& 19.90 & \underline{0.660} & \underline{0.292}
& 21.11 & \underline{0.687} & \underline{0.259} \\

\midrule

DAVANet~\cite{zhou2019davanet}
& \checkmark & \checkmark
& 19.34 & 0.573 & 0.360
& 19.61 & 0.625 & 0.292 \\

Difix3D+~\cite{wu2025difix3d+}
& \checkmark & \checkmark
& \underline{19.94} & 0.611 & 0.311
& \underline{21.30} & 0.673 & 0.262 \\

\textbf{Ours}
& \checkmark & \checkmark
& \textbf{21.13} & \textbf{0.700} & \textbf{0.228}
& \textbf{23.41} & \textbf{0.796} & \textbf{0.165} \\

\bottomrule
\end{tabular}%
}
\end{table}

\begin{figure}[!t]
    \centering
    \includegraphics[width=\textwidth]{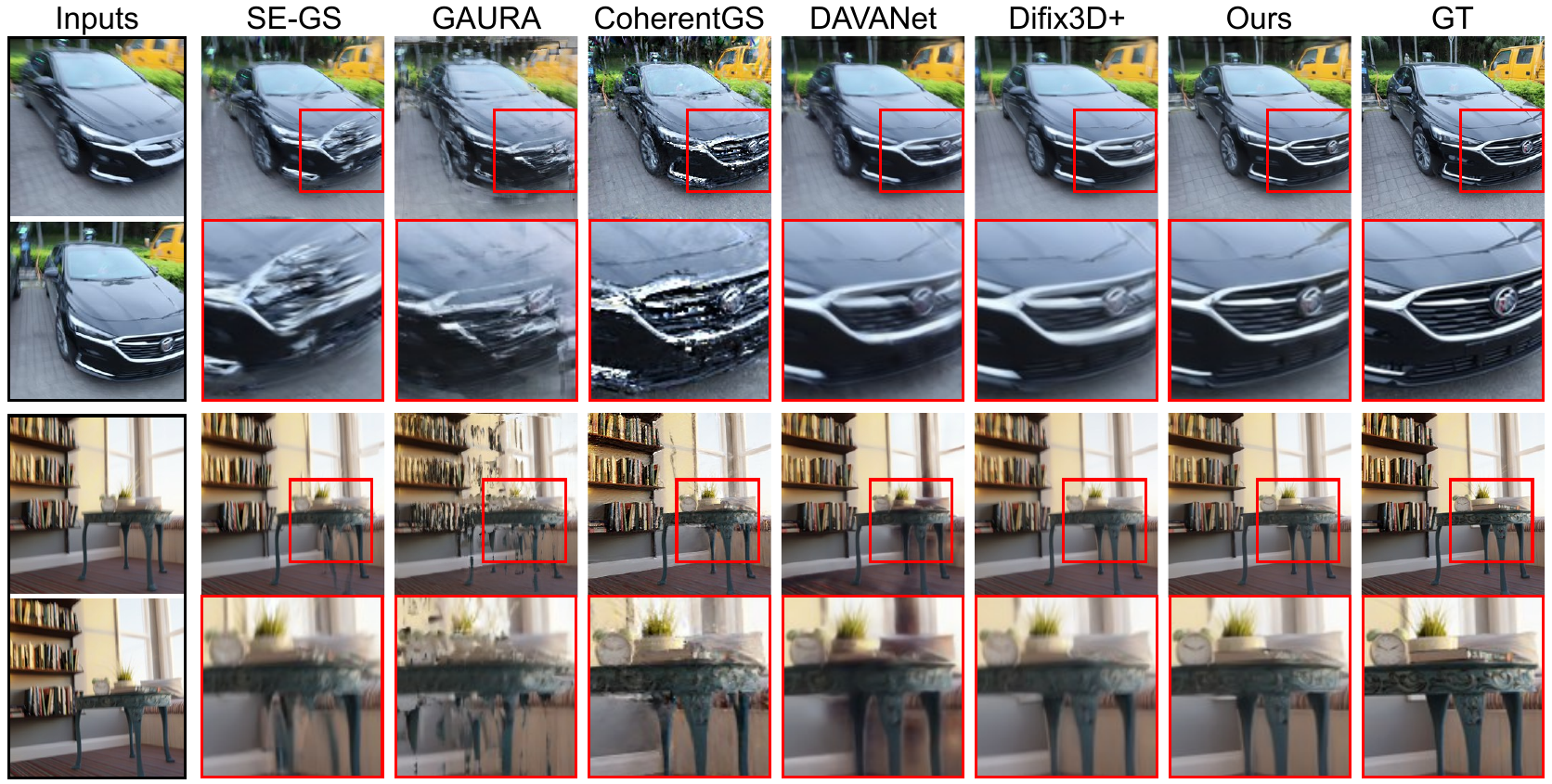}
    \caption{
        Qualitative novel-view synthesis comparisons on real-world and synthetic Deblur-NeRF scenes.
        The upper vehicle scene is real-world and the lower indoor scene synthetic.
        The Inputs column shows the two blurry views.
        For each method, the upper image shows the rendered target view, while the lower enlarges the red-boxed region.      
    }
    \label{fig:qualitative}
\end{figure}

\paragraph{\normalfont\bfseries Evaluation metrics.}
We evaluate novel-view renderings against the corresponding sharp reference images using PSNR, SSIM, and LPIPS.
We report the average over all fixed evaluation tuples separately for the synthetic and real-world subsets.
Higher PSNR and SSIM indicate better reconstruction quality, whereas lower LPIPS indicates greater perceptual similarity.

\subsection{Comparison with Prior Methods}
\label{subsec:comparison}
\paragraph{\normalfont\bfseries Quantitative results.}
\Cref{tab:main_results} compares CasDeblurGS with prior methods on the real-world and synthetic Deblur-NeRF subsets.
Our method achieves the best performance across all metrics on both subsets.
On real-world scenes, CasDeblurGS obtains 21.13 dB PSNR, 0.700 SSIM, and 0.228 LPIPS, outperforming the best competing results by 1.19 dB, 0.040, and 0.064, respectively.
The gains are larger on synthetic scenes, where our method improves PSNR by 2.11 dB, SSIM by 0.109, and LPIPS by 0.094.

The gaps also reflect differences between the baselines' native settings and our extreme two-view regime.
SE-GS addresses few-shot reconstruction from sparse posed views but does not explicitly handle blur-corrupted correspondences.
GAURA is trained with 8--12 source views and uses 10 views at inference, leaving substantially less multi-view redundancy for epipolar aggregation when restricted to two inputs.
CoherentGS directly targets sparse motion blur but reports configurations with 3, 6, or 9 views, leaving its alternating reconstruction and generative expansion more weakly constrained under only two inputs.

The consistent gains over DAVANet and Difix3D+, which use the same frozen NoPoSplat backbone, show that applying a strong 2D restoration model before reconstruction is insufficient.
Instead, the results demonstrate the benefit of progressively incorporating reliable local correspondence cues and global 3D guidance.
Moreover, CasDeblurGS remains pose-free and generalizable without per-scene optimization.

\paragraph{\normalfont\bfseries Qualitative results.}
As shown in \cref{fig:qualitative}, SE-GS and GAURA retain substantial blur, while CoherentGS and the restoration-based baselines partially reduce blur but oversmooth details or introduce structural artifacts.
These limitations are particularly visible around the vehicle grille in the real-world example and the table edges and legs in the synthetic example.
In contrast, CasDeblurGS recovers sharper boundaries and structures that more closely match the ground-truth views.
These results show that the cascade improves both sharpness and structural fidelity in novel-view renderings.
Additional qualitative results are provided in the supplementary material.

\subsection{Ablation Studies}
\label{subsec:ablation}

\paragraph{\normalfont\bfseries Progressive contribution of the cascade.}
\begin{wraptable}{r}{0.46\textwidth}
\vspace{-1.19cm}
\centering
\caption{Ablation on real-world scenes.}
\label{tab:progressive_ablation}
\small
\setlength{\tabcolsep}{5pt}
\renewcommand{\arraystretch}{1.0}
\resizebox{\linewidth}{!}{%
\begin{tabular}{@{}lcccc@{}}
\toprule
& Blurry & +Stab. & +2D & +3D \\
\midrule
PSNR $\uparrow$
& 20.07 & 20.30 & \underline{20.88} & \textbf{21.13} \\
SSIM $\uparrow$
& 0.612 & 0.655 & \underline{0.684} & \textbf{0.700} \\
\bottomrule
\end{tabular}%
}
\vspace{-0.8cm}
\end{wraptable}
\Cref{tab:progressive_ablation} shows consistent gains as the stabilizer, Stage~1 2D guidance, and Stage~2 3D guidance are progressively introduced.
The stabilizer provides an initial improvement by producing more alignment-friendly observations.
Stage~1 yields the largest incremental PSNR gain ($+0.58$ dB), confirming the benefit of locally reliable cross-view correspondences.
Stage~2 further improves PSNR by $0.25$ dB and SSIM by $0.016$, indicating that dense global guidance resolves residual inconsistencies that cannot be addressed by 2D warping alone.
Additional component diagnostics are provided in the supplementary material.

\begin{figure}[!b]
    \centering
    \includegraphics[width=\textwidth]{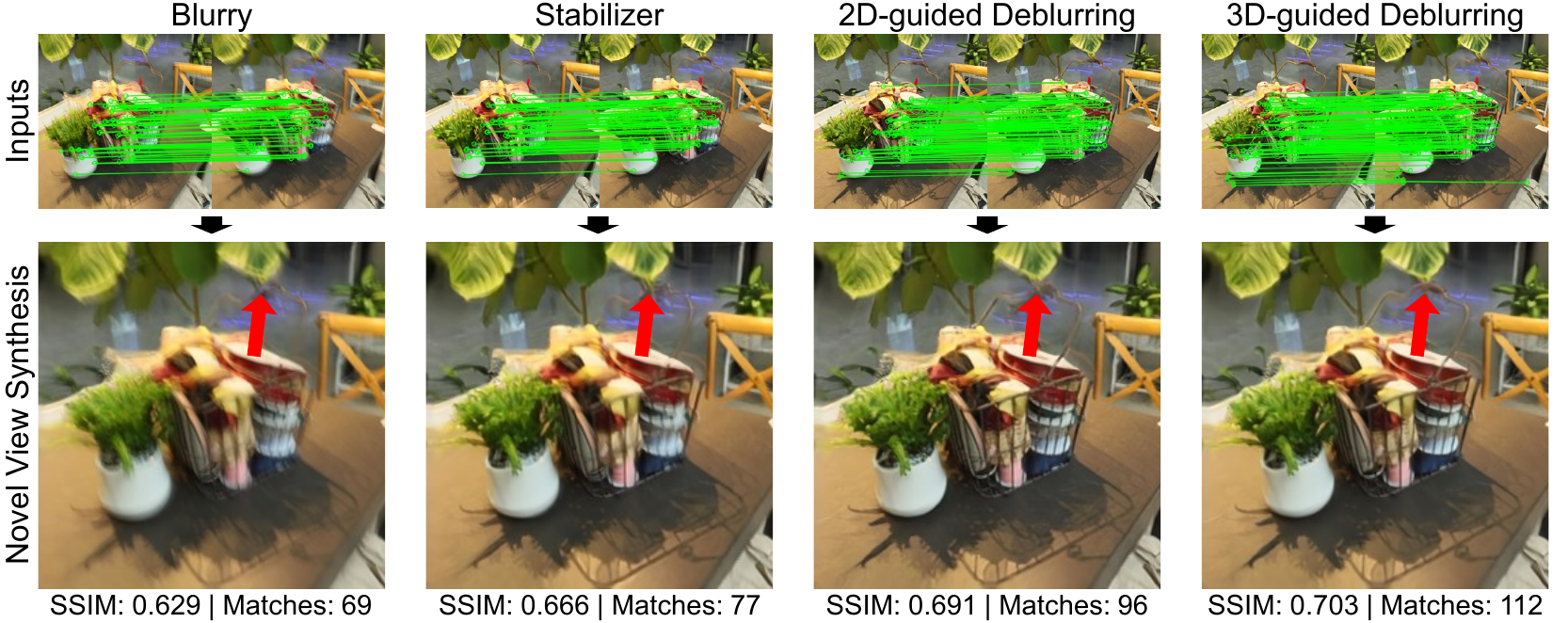}
    \caption{
        Progressive ablation on real-world BlurBasket scene. 
        Green lines denote cross-view 2D matches between input views, and red arrows highlight the basket handle.
    }
    \label{fig:qualitative_ablation}
\end{figure}

\paragraph{\normalfont\bfseries Cross-view correspondence analysis.}
\Cref{fig:qualitative_ablation} visualizes the progressive improvement in cross-view consistency.
On BlurBasket, the number of matches increases from 69 for the blurry inputs to 77 after stabilization, 96 after Stage~1, and 112 after Stage~2.
The limited gain from stabilization reflects its independent per-view processing, whereas Stage~1 establishes more reliable local correspondences through OCGM.
Stage~2 further improves correspondence recovery in regions that remain challenging for 2D warping alone, such as the basket handle highlighted by the red arrows.
These results show that the cascade progressively improves both local correspondence reliability and global structural consistency.

\paragraph{\normalfont\bfseries Camera reprojection analysis.}
To examine whether Stage~2 improves cross-view geometric consistency rather than merely image sharpness, we conduct the reprojection analysis reported in \cref{tab:camera_reprojection}.
For each restored input pair, we extract mutual SIFT matches, triangulate them using the benchmark camera projection matrices, and compute symmetric reprojection errors in both views.
We report averages across the evaluation pairs for valid triangulated points, median reprojection error, one-pixel inliers, and inlier ratio.
The benchmark camera poses are used only for this evaluation and are never provided to our pose-free reconstruction pipeline.
Compared with Stage~1, Stage~2 produces approximately 33\% more valid triangulated points and one-pixel inliers while reducing reprojection error by 9.7\%.
The inlier ratio increases from 0.9168 to 0.9343, indicating that 3D re-render guidance improves the geometric consistency of the restored views.

\begin{table}[t]
\caption{Camera reprojection analysis on real-world scenes.}
\label{tab:camera_reprojection}
\centering
\small
\setlength{\tabcolsep}{3pt}
\renewcommand{\arraystretch}{1.0}
\begin{tabular*}{\columnwidth}{@{\extracolsep{\fill}}lcccc@{}}
\toprule
Stage
& Valid pts $\uparrow$
& Reproj. err. $\downarrow$
& Inliers@1px $\uparrow$
& Inlier ratio $\uparrow$ \\
\midrule
Stage~1
& 95.1
& 0.2215
& 85.4
& 0.9168 \\
Stage~2
& \textbf{126.3}
& \textbf{0.2001}
& \textbf{113.5}
& \textbf{0.9343} \\
\bottomrule
\end{tabular*}
\end{table}

\section{Conclusion}
\label{sec:conclusion}
We presented CasDeblurGS, a cascaded framework for pose-free 3D Gaussian Splatting from two motion-blurred images with known camera intrinsics.
Our method progressively recovers reliable cross-view information: Stage~1 constructs locally trustworthy 2D guidance via occlusion-aware correspondence filtering, while Stage~2 uses pose-free 3D re-rendering to provide dense global guidance for final restoration.
The resulting views enable coherent 3D reconstruction with a frozen feed-forward 3DGS backbone without input-view poses, auxiliary sharp images, or per-scene test-time optimization.
Experiments on real-world and synthetic Deblur-NeRF scenes demonstrate consistent gains over sparse-view reconstruction and restoration baselines.
Progressive ablations, reprojection analysis, and cross-view correspondence visualization further show that the proposed cascade improves both rendering quality and multi-view geometric consistency.

\paragraph{\normalfont\bfseries Limitations and future work.}
Our framework assumes known intrinsics and static scenes in a two-view setting.
Its correspondence guidance may deteriorate under extreme blur, large occlusions, or limited overlap.
Future work will address unknown intrinsics, dynamic scenes, and broader sparse-view settings.

\section*{Acknowledgements}
This work was supported by Institute of Information \& communications Technology Planning \& Evaluation (IITP) grant funded by the Korea government (MSIT) (No. RS-2026-25522885, Development of a World Foundation Model for Training and Deployment of Physical AI Systems).

\bibliographystyle{splncs04}
\bibliography{main}

@String(CVPR  = {IEEE Conf. Comput. Vis. Pattern Recog.})

@String(TOG   = {ACM Trans. Graph.})

@String(CVPR  = {CVPR})

@String(TOG   = {ACM TOG})

@inproceedings{mildenhall2020nerf,
  title={Nerf: Representing scenes as neural radiance fields for view synthesis},
  author={Mildenhall, B and Srinivasan, PP and Tancik, M and Barron, JT and Ramamoorthi, R and Ng, R},
  booktitle={European conference on computer vision},
  year={2020}
}

@article{kerbl2023gaussiansplatting,
  title={3d gaussian splatting for real-time radiance field rendering},
  author={Kerbl, Bernhard and Kopanas, Georgios and Leimk{\"u}hler, Thomas and Drettakis, George},
  journal={ACM Transactions on Graphics},
  volume={42},
  number={4},
  pages={1--14},
  year={2023},
  publisher={ACM}
}

@inproceedings{ma2022deblurnerf,
  title={Deblur-nerf: Neural radiance fields from blurry images},
  author={Ma, Li and Li, Xiaoyu and Liao, Jing and Zhang, Qi and Wang, Xuan and Wang, Jue and Sander, Pedro V},
  booktitle={Proceedings of the IEEE/CVF Conference on Computer Vision and Pattern Recognition},
  pages={12861--12870},
  year={2022}
}

@inproceedings{wang2023bad,
  title={Bad-nerf: Bundle adjusted deblur neural radiance fields},
  author={Wang, Peng and Zhao, Lingzhe and Ma, Ruijie and Liu, Peidong},
  booktitle={Proceedings of the IEEE/CVF Conference on Computer Vision and Pattern Recognition},
  pages={4170--4179},
  year={2023}
}

@inproceedings{lee2023dpnerf,
  title={Dp-nerf: Deblurred neural radiance field with physical scene priors},
  author={Lee, Dogyoon and Lee, Minhyeok and Shin, Chajin and Lee, Sangyoun},
  booktitle={Proceedings of the IEEE/CVF Conference on Computer Vision and Pattern Recognition},
  pages={12386--12396},
  year={2023}
}

@inproceedings{lee2023exblurf,
  title={ExBluRF: Efficient Radiance Fields for Extreme Motion Blurred Images},
  author={Lee, Dongwoo and Oh, Jeongtaek and Rim, Jaesung and Cho, Sunghyun and Lee, Kyoung Mu},
  booktitle={Proceedings of the IEEE/CVF International Conference on Computer Vision},
  pages={17639--17648},
  year={2023}
}

@inproceedings{lee2024deblurring,
  title={Deblurring 3d gaussian splatting},
  author={Lee, Byeonghyeon and Lee, Howoong and Sun, Xiangyu and Ali, Usman and Park, Eunbyung},
  booktitle={European Conference on Computer Vision},
  pages={127--143},
  year={2024},
  organization={Springer}
}

@inproceedings{peng2024bags,
  title={Bags: Blur agnostic gaussian splatting through multi-scale kernel modeling},
  author={Peng, Cheng and Tang, Yutao and Zhou, Yifan and Wang, Nengyu and Liu, Xijun and Li, Deming and Chellappa, Rama},
  booktitle={European Conference on Computer Vision},
  pages={293--310},
  year={2024},
  organization={Springer}
}

@inproceedings{lee2025comogaussian,
  title={Comogaussian: Continuous motion-aware gaussian splatting from motion-blurred images},
  author={Lee, Jungho and Kim, Donghyeong and Lee, Dogyoon and Cho, Suhwan and Lee, Minhyeok and Lee, Wonjoon and Kim, Taeoh and Wee, Dongyoon and Lee, Sangyoun},
  booktitle={Proceedings of the IEEE/CVF International Conference on Computer Vision},
  pages={26415--26424},
  year={2025}
}

@inproceedings{choi2025exploiting,
  title={Exploiting deblurring networks for radiance fields},
  author={Choi, Haeyun and Yang, Heemin and Han, Janghyeok and Cho, Sunghyun},
  booktitle={Proceedings of the Computer Vision and Pattern Recognition Conference},
  pages={6012--6021},
  year={2025}
}

@inproceedings{zhao2024bad,
  title={Bad-gaussians: Bundle adjusted deblur gaussian splatting},
  author={Zhao, Lingzhe and Wang, Peng and Liu, Peidong},
  booktitle={European Conference on Computer Vision},
  pages={233--250},
  year={2024},
  organization={Springer}
}

@inproceedings{zhao2025bsgs,
  title={BSGS: Bi-stage 3D Gaussian Splatting for Camera Motion Deblurring},
  author={Zhao, An and Yu, Piaopiao and Zhu, Zhe and Wei, Mingqiang},
  booktitle={Proceedings of the 33rd ACM International Conference on Multimedia},
  pages={8351--8359},
  year={2025}
}

@article{lee2025sparse,
  title={Sparse-DeRF: Deblurred neural radiance fields from sparse view},
  author={Lee, Dogyoon and Kim, Donghyeong and Lee, Jungho and Lee, Minhyeok and Lee, Seunghoon and Lee, Sangyoun},
  journal={IEEE Transactions on Pattern Analysis and Machine Intelligence},
  year={2025},
  publisher={IEEE}
}

@article{xu2025breaking,
  title={Breaking the Vicious Cycle: Coherent 3D Gaussian Splatting from Sparse and Motion-Blurred Views},
  author={Xu, Zhankuo and Feng, Chaoran and Li, Yingtao and Zhao, Jianbin and Yang, Jiashu and Yu, Wangbo and Yuan, Li and Tian, Yonghong},
  journal={arXiv preprint arXiv:2512.10369},
  year={2025}
}

@inproceedings{charatan2024pixelsplat,
  title={pixelsplat: 3d gaussian splats from image pairs for scalable generalizable 3d reconstruction},
  author={Charatan, David and Li, Sizhe Lester and Tagliasacchi, Andrea and Sitzmann, Vincent},
  booktitle={Proceedings of the IEEE/CVF conference on computer vision and pattern recognition},
  pages={19457--19467},
  year={2024}
}

@inproceedings{chen2024mvsplat,
  title={Mvsplat: Efficient 3d gaussian splatting from sparse multi-view images},
  author={Chen, Yuedong and Xu, Haofei and Zheng, Chuanxia and Zhuang, Bohan and Pollefeys, Marc and Geiger, Andreas and Cham, Tat-Jen and Cai, Jianfei},
  booktitle={European conference on computer vision},
  pages={370--386},
  year={2024},
  organization={Springer}
}

@inproceedings{xu2025depthsplat,
  title={Depthsplat: Connecting gaussian splatting and depth},
  author={Xu, Haofei and Peng, Songyou and Wang, Fangjinhua and Blum, Hermann and Barath, Daniel and Geiger, Andreas and Pollefeys, Marc},
  booktitle={Proceedings of the Computer Vision and Pattern Recognition Conference},
  pages={16453--16463},
  year={2025}
}

@inproceedings{ye2024no,
  title     = {No Pose, No Problem: Surprisingly Simple {3D} Gaussian Splats from Sparse Unposed Images},
  author    = {Ye, Botao and Liu, Sifei and Xu, Haofei and Li, Xueting and
               Pollefeys, Marc and Yang, Ming-Hsuan and Peng, Songyou},
  booktitle = {International Conference on Learning Representations},
  year      = {2025}
}

@article{jiang2025anysplat,
  title={Anysplat: Feed-forward 3d gaussian splatting from unconstrained views},
  author={Jiang, Lihan and Mao, Yucheng and Xu, Linning and Lu, Tao and Ren, Kerui and Jin, Yichen and Xu, Xudong and Yu, Mulin and Pang, Jiangmiao and Zhao, Feng and others},
  journal={ACM Transactions on Graphics (TOG)},
  volume={44},
  number={6},
  pages={1--16},
  year={2025},
  publisher={ACM New York, NY, USA}
}

@article{hong2024pf3plat,
  title={Pf3plat: Pose-free feed-forward 3d gaussian splatting},
  author={Hong, Sunghwan and Jung, Jaewoo and Shin, Heeseong and Han, Jisang and Yang, Jiaolong and Luo, Chong and Kim, Seungryong},
  journal={arXiv preprint arXiv:2410.22128},
  year={2024}
}

@inproceedings{chen2022simple,
  title={Simple baselines for image restoration},
  author={Chen, Liangyu and Chu, Xiaojie and Zhang, Xiangyu and Sun, Jian},
  booktitle={European conference on computer vision},
  pages={17--33},
  year={2022},
  organization={Springer}
}

@inproceedings{teed2020raft,
  title={Raft: Recurrent all-pairs field transforms for optical flow},
  author={Teed, Zachary and Deng, Jia},
  booktitle={European conference on computer vision},
  pages={402--419},
  year={2020},
  organization={Springer}
}

@inproceedings{wang2021realesrgan,
    author = {Xintao Wang and Liangbin Xie and Chao Dong and Ying Shan},
    title = {Real-ESRGAN: Training Real-World Blind Super-Resolution with Pure Synthetic Data},
    booktitle = {International Conference on Computer Vision Workshops (ICCVW)},
    year = {2021}
}

@article{simonyan2014very,
  title={Very deep convolutional networks for large-scale image recognition},
  author={Simonyan, Karen and Zisserman, Andrew},
  journal={arXiv preprint arXiv:1409.1556},
  year={2014}
}

@InProceedings{Nah_2017_CVPR,
  author = {Nah, Seungjun and Kim, Tae Hyun and Lee, Kyoung Mu},
  title = {Deep Multi-Scale Convolutional Neural Network for Dynamic Scene Deblurring},
  booktitle = {CVPR},
  month = {July},
  year = {2017}
}

@misc{torchvision2016,
  author       = {{TorchVision maintainers and contributors}},
  title        = {{TorchVision}: {PyTorch}'s Computer Vision Library},
  year         = {2016},
  howpublished = {\url{https://github.com/pytorch/vision}}
}

@article{bao20253d,
  title={3d gaussian splatting: Survey, technologies, challenges, and opportunities},
  author={Bao, Yanqi and Ding, Tianyu and Huo, Jing and Liu, Yaoli and Li, Yuxin and Li, Wenbin and Gao, Yang and Luo, Jiebo},
  journal={IEEE Transactions on Circuits and Systems for Video Technology},
  volume={35},
  number={7},
  pages={6832--6852},
  year={2025},
  publisher={IEEE}
}

@article{joshi2025unconstrained,
  title={Unconstrained large-scale 3d reconstruction and rendering across altitudes},
  author={Joshi, Neil and Carney, Joshua and Kuo, Nathanael and Li, Homer and Peng, Cheng and Brown, Myron},
  journal={arXiv preprint arXiv:2505.00734},
  year={2025}
}

@article{gao2024cat3d,
  title={Cat3d: Create anything in 3d with multi-view diffusion models},
  author={Gao, Ruiqi and Holynski, Aleksander and Henzler, Philipp and Brussee, Arthur and Martin-Brualla, Ricardo and Srinivasan, Pratul and Barron, Jonathan T and Poole, Ben},
  journal={arXiv preprint arXiv:2405.10314},
  year={2024}
}

@inproceedings{khan2025autosplat,
  title={Autosplat: Constrained gaussian splatting for autonomous driving scene reconstruction},
  author={Khan, Mustafa and Fazlali, Hamidreza and Sharma, Dhruv and Cao, Tongtong and Bai, Dongfeng and Ren, Yuan and Liu, Bingbing},
  booktitle={2025 IEEE International Conference on Robotics and Automation (ICRA)},
  pages={8315--8321},
  year={2025},
  organization={IEEE}
}

@inproceedings{zhang2025egogaussian,
  title={Egogaussian: Dynamic scene understanding from egocentric video with 3d gaussian splatting},
  author={Zhang, Daiwei and Li, Gengyan and Li, Jiajie and Bressieux, Micka{\"e}l and Hilliges, Otmar and Pollefeys, Marc and Van Gool, Luc and Wang, Xi},
  booktitle={2025 International Conference on 3D Vision (3DV)},
  pages={1091--1102},
  year={2025},
  organization={IEEE}
}

@inproceedings{schonberger2016structure,
  title={Structure-from-motion revisited},
  author={Schonberger, Johannes L and Frahm, Jan-Michael},
  booktitle={Proceedings of the IEEE conference on computer vision and pattern recognition},
  pages={4104--4113},
  year={2016}
}

@article{lee2024crim,
  title={CRiM-GS: Continuous Rigid Motion-Aware Gaussian Splatting from Motion-Blurred Images},
  author={Lee, Jungho and Kim, Donghyeong and Lee, Dogyoon and Cho, Suhwan and Lee, Minhyeok and Lee, Sangyoun},
  journal={arXiv preprint arXiv:2407.03923},
  year={2024}
}

@article{wu2024deblur4dgs,
  title={Deblur4dgs: 4d gaussian splatting from blurry monocular video},
  author={Wu, Renlong and Zhang, Zhilu and Chen, Mingyang and Yan, Zifei and Zuo, Wangmeng},
  journal={arXiv preprint arXiv:2412.06424},
  year={2024}
}

@inproceedings{gupta2024gaura,
  title={GAURA: Generalizable Approach for Unified Restoration and Rendering of Arbitrary Views},
  author={Gupta, Vinayak and Girish, Rongali Simhachala Venkata and Mukund Varma, T and Tewari, Ayush and Mitra, Kaushik},
  booktitle={European Conference on Computer Vision},
  pages={249--266},
  year={2024},
  organization={Springer}
}

@inproceedings{lin2025hqgs,
  title={Hqgs: High-quality novel view synthesis with gaussian splatting in degraded scenes},
  author={Lin, Xin and Luo, Shi and Shan, Xiaojun and Zhou, Xiaoyu and Ren, Chao and Qi, Lu and Yang, Ming-Hsuan and Vasconcelos, Nuno},
  booktitle={The Thirteenth International Conference on Learning Representations},
  year={2025}
}

@inproceedings{wan2025s2gaussian,
  title={S2gaussian: Sparse-view super-resolution 3d gaussian splatting},
  author={Wan, Yecong and Shao, Mingwen and Cheng, Yuanshuo and Zuo, Wangmeng},
  booktitle={Proceedings of the Computer Vision and Pattern Recognition Conference},
  pages={711--721},
  year={2025}
}

@inproceedings{zhang2025pansplat,
  title={Pansplat: 4k panorama synthesis with feed-forward gaussian splatting},
  author={Zhang, Cheng and Xu, Haofei and Wu, Qianyi and Gambardella, Camilo Cruz and Phung, Dinh and Cai, Jianfei},
  booktitle={Proceedings of the Computer Vision and Pattern Recognition Conference},
  pages={11437--11447},
  year={2025}
}

@inproceedings{wei2025omni,
  title={Omni-scene: Omni-gaussian representation for ego-centric sparse-view scene reconstruction},
  author={Wei, Dongxu and Li, Zhiqi and Liu, Peidong},
  booktitle={Proceedings of the Computer Vision and Pattern Recognition Conference},
  pages={22317--22327},
  year={2025}
}

@inproceedings{zhou2019davanet,
  title={Davanet: Stereo deblurring with view aggregation},
  author={Zhou, Shangchen and Zhang, Jiawei and Zuo, Wangmeng and Xie, Haozhe and Pan, Jinshan and Ren, Jimmy S},
  booktitle={Proceedings of the IEEE/CVF Conference on Computer Vision and Pattern Recognition},
  pages={10996--11005},
  year={2019}
}

@inproceedings{yan2020disparity,
  title={Disparity-aware domain adaptation in stereo image restoration},
  author={Yan, Bo and Ma, Chenxi and Bare, Bahetiyaer and Tan, Weimin and Hoi, Steven CH},
  booktitle={Proceedings of the IEEE/CVF Conference on Computer Vision and Pattern Recognition},
  pages={13179--13187},
  year={2020}
}

@inproceedings{pan2017simultaneous,
  title={Simultaneous stereo video deblurring and scene flow estimation},
  author={Pan, Liyuan and Dai, Yuchao and Liu, Miaomiao and Porikli, Fatih},
  booktitle={Proceedings of the IEEE conference on computer vision and pattern recognition},
  pages={4382--4391},
  year={2017}
}

@inproceedings{wu2025difix3d+,
  title={Difix3d+: Improving 3d reconstructions with single-step diffusion models},
  author={Wu, Jay Zhangjie and Zhang, Yuxuan and Turki, Haithem and Ren, Xuanchi and Gao, Jun and Shou, Mike Zheng and Fidler, Sanja and Gojcic, Zan and Ling, Huan},
  booktitle={Proceedings of the IEEE/CVF Conference on Computer Vision and Pattern Recognition},
  pages={26024--26035},
  year={2025}
}

@inproceedings{mao2025sir,
  title={Sir-diff: Sparse image sets restoration with multi-view diffusion model},
  author={Mao, Yucheng and Wang, Boyang and Kulkarni, Nilesh and Park, Jeong Joon},
  booktitle={Proceedings of the Computer Vision and Pattern Recognition Conference},
  pages={21620--21630},
  year={2025}
}

@inproceedings{luo20253denhancer,
  title={3denhancer: Consistent multi-view diffusion for 3d enhancement},
  author={Luo, Yihang and Zhou, Shangchen and Lan, Yushi and Pan, Xingang and Loy, Chen Change},
  booktitle={Proceedings of the Computer Vision and Pattern Recognition Conference},
  pages={16430--16440},
  year={2025}
}

@inproceedings{yu2021pixelnerf,
  title={pixelnerf: Neural radiance fields from one or few images},
  author={Yu, Alex and Ye, Vickie and Tancik, Matthew and Kanazawa, Angjoo},
  booktitle={Proceedings of the IEEE/CVF conference on computer vision and pattern recognition},
  pages={4578--4587},
  year={2021}
}

@inproceedings{wang2021ibrnet,
  title={Ibrnet: Learning multi-view image-based rendering},
  author={Wang, Qianqian and Wang, Zhicheng and Genova, Kyle and Srinivasan, Pratul P and Zhou, Howard and Barron, Jonathan T and Martin-Brualla, Ricardo and Snavely, Noah and Funkhouser, Thomas},
  booktitle={Proceedings of the IEEE/CVF conference on computer vision and pattern recognition},
  pages={4690--4699},
  year={2021}
}

@inproceedings{chen2021mvsnerf,
  title={Mvsnerf: Fast generalizable radiance field reconstruction from multi-view stereo},
  author={Chen, Anpei and Xu, Zexiang and Zhao, Fuqiang and Zhang, Xiaoshuai and Xiang, Fanbo and Yu, Jingyi and Su, Hao},
  booktitle={Proceedings of the IEEE/CVF international conference on computer vision},
  pages={14124--14133},
  year={2021}
}

@article{zhang2025advances,
  title={Advances in feed-forward 3d reconstruction and view synthesis: A survey},
  author={Zhang, Jiahui and Li, Yuelei and Chen, Anpei and Xu, Muyu and Liu, Kunhao and Wang, Jianyuan and Long, Xiao-Xiao and Liang, Hanxue and Xu, Zexiang and Su, Hao and others},
  journal={arXiv preprint arXiv:2507.14501},
  year={2025}
}

@article{tang2024hisplat,
  title={Hisplat: Hierarchical 3d gaussian splatting for generalizable sparse-view reconstruction},
  author={Tang, Shengji and Ye, Weicai and Ye, Peng and Lin, Weihao and Zhou, Yang and Chen, Tao and Ouyang, Wanli},
  journal={arXiv preprint arXiv:2410.06245},
  year={2024}
}

@inproceedings{zhao2025self,
  title={Self-ensembling gaussian splatting for few-shot novel view synthesis},
  author={Zhao, Chen and Wang, Xuan and Zhang, Tong and Javed, Saqib and Salzmann, Mathieu},
  booktitle={Proceedings of the IEEE/CVF International Conference on Computer Vision},
  pages={4940--4950},
  year={2025}
}
\end{document}


\title{
Supplementary Material:
CasDeblurGS: Cascaded 2D-to-3D Multi-View Consistency for 3D Gaussian Splatting from Two Blurry Images
}

\titlerunning{Supplementary Material for CasDeblurGS}

\author{
Haeyun Choi\inst{1}\textsuperscript{*}\textsuperscript{$\dagger$}
\orcidlink{0009-0006-6399-4331}
\and
Minhyuk Jang\inst{2}\textsuperscript{*}
\orcidlink{0009-0001-5549-3864}
\and
I-Gil Kim\inst{2}
\orcidlink{0009-0001-4938-0038}
}

\authorrunning{H.~Choi et al.}

\institute{
University of Virginia, Charlottesville, VA, USA\\
\email{phh3ps@virginia.edu}
\and
KT R\&D Center, Seoul, Republic of Korea\\
\email{\{minhyuk.jang,i-gil.kim\}@kt.com}
}

\maketitle

\begingroup
\renewcommand{\thefootnote}{*}
\footnotetext{Equal contribution.}

\renewcommand{\thefootnote}{\ensuremath{\dagger}}
\footnotetext{This work was done at KT R\&D Center.}
\endgroup

\vspace{1.5em}
\begin{center}
    {\large\bfseries Supplementary Contents}
\end{center}
\vspace{-0.5em}

\setcounter{tocdepth}{2}

\makeatletter
\begingroup
\let\l@title\@gobbletwo
\let\l@author\@gobbletwo
\def\authcount#1{}
\@starttoc{toc}
\endgroup
\makeatother

\clearpage

\section{Additional Implementation Details}
\label{sec:additional_impl}

In this section, we provide additional implementation details for reproducibility, including dataset construction, frozen modules, stage-wise training, and baseline adaptation.

\subsection{Training and Evaluation Data}
For training and validation, we use the synthetic camera-motion-blur dataset introduced in DeepDeblurRF~\cite{choi2025exploiting}. 
The training split contains 65 scenes, each with 29 viewpoints. 
For each viewpoint, one blurry image and its corresponding sharp target are provided. 
The validation split contains 10 additional scenes with the same structure.
Blurred images are generated by simulating 6-DoF camera motion during exposure: multiple intermediate views are rendered along a smooth camera trajectory, averaged in linear color space, and paired with the temporally central frame as the ground-truth sharp image.

For evaluation, we use the camera-motion-blur subset of Deblur-NeRF~\cite{ma2022deblurnerf}, which contains five synthetic scenes and ten real-world scenes. 
Each scene provides motion-blurred input images and blur-free ground-truth target views for evaluation.
To ensure fair and reproducible comparison in the extreme two-view setting, we use a fixed set of per-scene evaluation instances, where each instance is defined by two blurry input views and one held-out target view for novel view synthesis. 
The complete index mappings are provided in \cref{tab:transposed_indices_final}, and all compared methods are evaluated using the same instance definitions.
For both training and evaluation, all images are first resized while preserving aspect ratio and then center-cropped to $256\times256$, following PixelSplat-style preprocessing~\cite{charatan2024pixelsplat}.

\subsection{Frozen Modules and OCGM Implementation}
We use the width-64 NAFNet~\cite{chen2022simple} pretrained on GoPro~\cite{Nah_2017_CVPR} as the frozen stabilizer $\mathcal{S}_{\phi}$, and RAFT-Large~\cite{teed2020raft} from Torchvision~\cite{torchvision2016} for optical flow estimation in OCGM. 
For each pair of stabilized views, we estimate bidirectional optical flow and compute validity masks using forward--backward consistency together with out-of-bounds checks. 
A pixel is marked invalid if its mapped coordinate falls outside the image boundary or if the forward--backward residual exceeds the motion-adaptive threshold defined in the main paper. 
These masks suppress unreliable regions caused by occlusions, motion boundaries, and blur-induced mismatches before masked warps are constructed for Stage~1 guidance. 
Invalid regions are zeroed out in the warped reference image, and the resulting masked warp together with the validity mask are fed to the Stage~1 deblurring network. 
This results in more reliable Stage~1 guidance by removing unstable correspondences near occlusions and motion boundaries.
The same bidirectional flow fields are used for both validity checking and masked warping.

\subsection{Frozen 3DGS Backbone and Input-View Re-Rendering}
We adopt the official NoPoSplat~\cite{ye2024no} checkpoint pretrained on RealEstate10K with two-view inputs at $256\times256$, and keep it frozen without further fine-tuning.
During Stage~2, the intermediate restorations and their camera intrinsics are fed into the backbone to construct a canonical-space 3D Gaussian representation. 
We then re-render this representation to the two input viewpoints to obtain the global 3D guidance images $R_1$ and $R_2$ for final restoration. 
These rendered RGB images are used directly as Stage~2 guidance.
Concretely, the first input view defines the canonical frame, and the second guidance image is rendered at the relative pose inferred by the backbone under the same canonical-space formulation.
No ground-truth input-view poses are used.
The same backbone is then applied to the restored views and their intrinsics to construct the output 3D representation used for novel view synthesis evaluation.

\subsection{Training Details}
Only the two deblurring networks, $\mathcal{D}^{2D}_{\theta}$ and $\mathcal{D}^{3D}_{\psi}$, are optimized during training. 
The stabilizer $\mathcal{S}_{\phi}$, RAFT, and the pose-free 3DGS backbone $h_{\eta}$ remain frozen throughout, with no gradients propagated through them.
Both trainable networks follow a NAFNet-style encoder--decoder architecture with base width 64. 
We use an encoder block configuration $[1,1,1,28]$, a single middle block, and a decoder configuration $[1,1,1,1]$. 
Stage~1 takes a 7-channel input formed by concatenating the 3-channel blurry base view, the 3-channel masked warp, and the 1-channel validity mask, while Stage~2 takes a 6-channel input formed by concatenating the 3-channel blurry base view and the 3-channel re-render guidance. 
Both networks output a restored 3-channel RGB image.

Training is performed in a stage-wise manner. 
In Stage~1, the frozen stabilizer and RAFT module are used to construct the masked warps and validity masks, and $\mathcal{D}^{2D}_{\theta}$ is trained to predict the intermediate restorations $(\tilde{S}_1,\tilde{S}_2)$ from the blurry inputs and OCGM guidance. 
After Stage~1 training, $\mathcal{D}^{2D}_{\theta}$ is frozen. 
In Stage~2, the frozen Stage~1 network first produces $(\tilde{S}_1,\tilde{S}_2)$, which are passed to the frozen pose-free 3DGS backbone to obtain the re-render guidance $(R_1,R_2)$. 
Using these re-rendered images, $\mathcal{D}^{3D}_{\psi}$ is trained to predict the final restorations $(\hat{S}_1,\hat{S}_2)$.

For each stage, we use AdamW with learning rate $1\times10^{-3}$, weight decay $1\times10^{-3}$, and $(\beta_1,\beta_2)=(0.9,0.9)$.
Each stage is trained for 400k iterations with cosine annealing to $\eta_{\min}=1\times10^{-7}$.
We use a batch size of 32 per GPU on two NVIDIA A100 80GB GPUs, with four dataloader workers per GPU. 
The training objective for both stages combines an $L_1$ reconstruction loss and a VGG19-based perceptual loss~\cite{simonyan2014very,wang2021realesrgan}:
\begin{equation}
\mathcal{L} = \lambda_1 \mathcal{L}_{\ell_1} + \lambda_p \mathcal{L}_{\text{perc}},
\end{equation}
where $\lambda_1=0.9$ and $\lambda_p=0.1$. No style loss is used.

\subsection{Baseline Adaptation Details}
All baselines are evaluated under the same fixed two-view protocol using the per-scene input--target index mappings listed in \cref{tab:transposed_indices_final}. 
Methods originally designed for different input regimes are adapted to this protocol while following their official implementations and recommended hyperparameter settings as closely as possible. 
For methods that require camera poses, we use the benchmark poses provided with Deblur-NeRF~\cite{ma2022deblurnerf}.

For SE-GS~\cite{zhao2025self}, we optimize each scene from the same two blurry source views defined by the protocol. 
For GAURA~\cite{gupta2024gaura}, we follow the official feed-forward inference pipeline in the two-view setting. 
For CoherentGS~\cite{xu2025breaking}, we retain its official alternating optimization framework while restricting the observations to the same two-view protocol.

For the control baselines DAVANet~\cite{zhou2019davanet} and Difix3D+~\cite{wu2025difix3d+}, the restored outputs are passed to the same frozen pose-free 3DGS backbone~\cite{ye2024no} used in our method. 
Thus, the ``Pose-Free'' and ``Generalizable'' labels reported in the main paper refer to the complete pipelines formed by each restoration model together with the shared frozen 3DGS backbone, rather than to the restoration models alone.



\begin{table}[tb]
\centering
\caption{Per-scene evaluation index mappings for the real-world and synthetic Deblur-NeRF scenes. Each instance is defined as $\{I_{blur}\}\rightarrow I_{tgt}$, where $\{I_{blur}\}$ denotes the two blurry input views and $I_{tgt}$ denotes the held-out target view for novel-view synthesis.}
\label{tab:transposed_indices_final}
\small 
\setlength{\tabcolsep}{5pt} 

\resizebox{\textwidth}{!}{ 
\begin{tabular}{l ccccccc}
\toprule
\textbf{Scene} & \textbf{Inst. 1} & \textbf{Inst. 2} & \textbf{Inst. 3} & \textbf{Inst. 4} & \textbf{Inst. 5} & \textbf{Inst. 6} & \textbf{Inst. 7} \\ \midrule

\multicolumn{8}{l}{\textit{Real-world Scenes}} \\ \midrule
BlurBall       & \{1, 2\} $\to$ 0  & \{6, 8\} $\to$ 7   & \{13, 15\} $\to$ 14 & \{19, 20\} $\to$ 21 & -- & -- & -- \\
BlurBasket     & \{1, 2\} $\to$ 0  & \{10, 11\} $\to$ 7 & \{18, 19\} $\to$ 14 & \{22, 23\} $\to$ 21 & \{27, 29\} $\to$ 28 & \{38, 39\} $\to$ 35 & \{41, 43\} $\to$ 42 \\
BlurBuick      & \{1, 2\} $\to$ 0  & \{10, 11\} $\to$ 7 & \{15, 18\} $\to$ 14 & \{4, 6\} $\to$ 21   & \{27, 33\} $\to$ 28 & \{36, 38\} $\to$ 35 & \{39, 43\} $\to$ 42 \\
BlurCoffee     & \{3, 8\} $\to$ 0  & \{10, 11\} $\to$ 6 & \{1, 17\} $\to$ 12  & \{5, 19\} $\to$ 18  & \{25, 26\} $\to$ 24 & -- & -- \\
BlurDecoration & \{1, 2\} $\to$ 0  & \{4, 9\} $\to$ 6   & \{3, 17\} $\to$ 12  & \{19, 33\} $\to$ 18 & \{16, 25\} $\to$ 24 & \{14, 29\} $\to$ 30 & \{31, 40\} $\to$ 36 \\
BlurGirl       & \{1, 2\} $\to$ 0  & \{6, 8\} $\to$ 7   & \{9, 11\} $\to$ 14  & \{12, 34\} $\to$ 21 & \{18, 29\} $\to$ 28 & \{30, 36\} $\to$ 35 & -- \\
BlurHeron      & \{1, 11\} $\to$ 0 & \{7, 15\} $\to$ 8  & \{15, 27\} $\to$ 16 & \{23, 25\} $\to$ 24 & \{19, 33\} $\to$ 32 & -- & -- \\
BlurParterre   & \{2, 9\} $\to$ 0  & \{4, 7\} $\to$ 6   & \{11, 13\} $\to$ 12 & \{19, 20\} $\to$ 18 & \{23, 27\} $\to$ 24 & \{1, 21\} $\to$ 30  & -- \\
BlurPuppet     & \{1, 2\} $\to$ 0  & \{10, 11\} $\to$ 6 & \{7, 8\} $\to$ 12   & \{15, 16\} $\to$ 18 & \{19, 21\} $\to$ 24 & \{27, 29\} $\to$ 30 & \{23, 37\} $\to$ 36 \\
BlurStair      & \{1, 2\} $\to$ 0  & \{7, 9\} $\to$ 6   & \{8, 17\} $\to$ 12  & \{21, 23\} $\to$ 18 & \{16, 26\} $\to$ 24 & \{22, 33\} $\to$ 30 & -- \\ \midrule

\multicolumn{8}{l}{\textit{Synthetic Scenes}} \\ \midrule
All Scenes        & \{1, 2\} $\to$ 0  & \{7, 9\} $\to$ 8   & \{15, 17\} $\to$ 16 & \{23, 25\} $\to$ 24 & \{31, 33\} $\to$ 32 & -- & -- \\ 
\bottomrule
\end{tabular}
}
\end{table}

\begin{figure}[t]
    \centering
    \includegraphics[width=\textwidth]{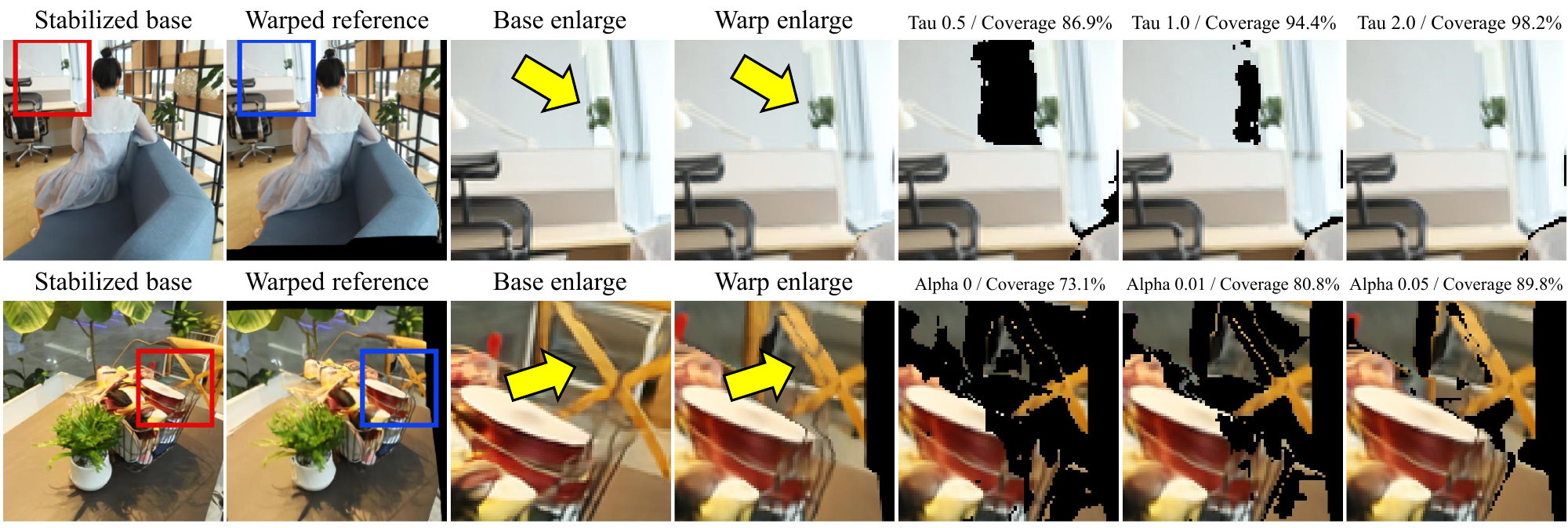}
    \caption{
    Qualitative effects of the OCGM thresholds.
    In each row, the stabilized base view $C_b$ provides the current-view
    context, while the unmasked warp
    $\mathcal{W}(C_r,F_{b\leftarrow r})$ transfers complementary content
    from the stabilized reference view $C_r$ into the base-view coordinates.
    The first row varies $\tau\in\{0.5,1.0,2.0\}$ with $\alpha=0.01$,
    whereas the second varies $\alpha\in\{0,0.01,0.05\}$ with $\tau=1.0$.
    Black regions in the masked guidance indicate rejected correspondences.
    The reported coverage denotes the retained fraction and is shown only
    to illustrate the trade-off between rejecting miswarped content and
    preserving useful cross-view information.
    }
    \label{fig:supp_ocgm_thresholds}
\end{figure}

\section{Qualitative Analysis of OCGM Thresholds}
\label{sec:supp_ocgm_thresholds}

Following the OCGM formulation in Eqs.~(8)--(10) of the main paper, we qualitatively examine how the minimum tolerance $\tau$ and the motion-dependent scale factor $\alpha$ affect the validity mask $M_{b\leftarrow r}$ and the masked cross-view warp $W_{b\leftarrow r}$ used as Stage~1 guidance.

For each example, $C_b$ denotes the stabilized base view, while $C_r$ denotes the stabilized reference view.
The unmasked warp $\mathcal{W}(C_r,F_{b\leftarrow r})$ transfers content from $C_r$ into the coordinates of $C_b$.
OCGM retains only the regions considered geometrically reliable and provides the resulting masked warp, together with its validity mask, to the Stage~1 restoration network.
The stabilizer output is therefore used only as an alignment-friendly observation: Stage~1 combines the original blurry base view with reliable complementary information from the reference view to produce a better restoration than independent stabilization alone.

For each threshold sweep, we keep $(C_b,C_r)$ and their bidirectional RAFT flows fixed while varying one parameter at a time.
Thus, the unmasked warp remains identical across the compared settings; only the regions retained as Stage~1 guidance change.

As shown in \cref{fig:supp_ocgm_thresholds}, the thresholds control the trade-off between rejecting unreliable warps and preserving valid cross-view information.
In the BlurGirl example, the unmasked warp incorrectly transfers the highlighted plant tip.
The permissive setting $\tau=2.0$ retains this miswarped structure, whereas $\tau=1.0$ rejects it while preserving most of the surrounding valid guidance.
The more conservative $\tau=0.5$ also removes the error but discards additional correctly warped content, as reflected by its lower
coverage.

The $\alpha$ sweep on BlurBasket shows a similar pattern in a large-displacement region.
The setting $\alpha=0.01$ selectively suppresses the distorted chair geometry while retaining much of the surrounding useful reference content.
In contrast, $\alpha=0$ rejects additional valid chair regions, whereas $\alpha=0.05$ admits distorted boundaries that should be excluded.
Together, these examples illustrate how overly conservative thresholds reduce the amount of usable guidance, while overly permissive thresholds propagate inaccurate reference content.

We use the fixed default $\tau=1.0$ and $\alpha=0.01$ for all training and evaluation scenes.
These values are not selected as per-scene optima; rather, they provide a practical scene-independent balance between filtering unreliable warps and retaining sufficient guidance across diverse scenes and motion patterns.
The alternative settings are shown only to illustrate OCGM's masking behavior and do not represent separately trained models.

\section{Additional Component Diagnostics}
\label{sec:supp_component_diagnostics}

\begin{figure}[t]
    \centering
    \includegraphics[width=\textwidth]{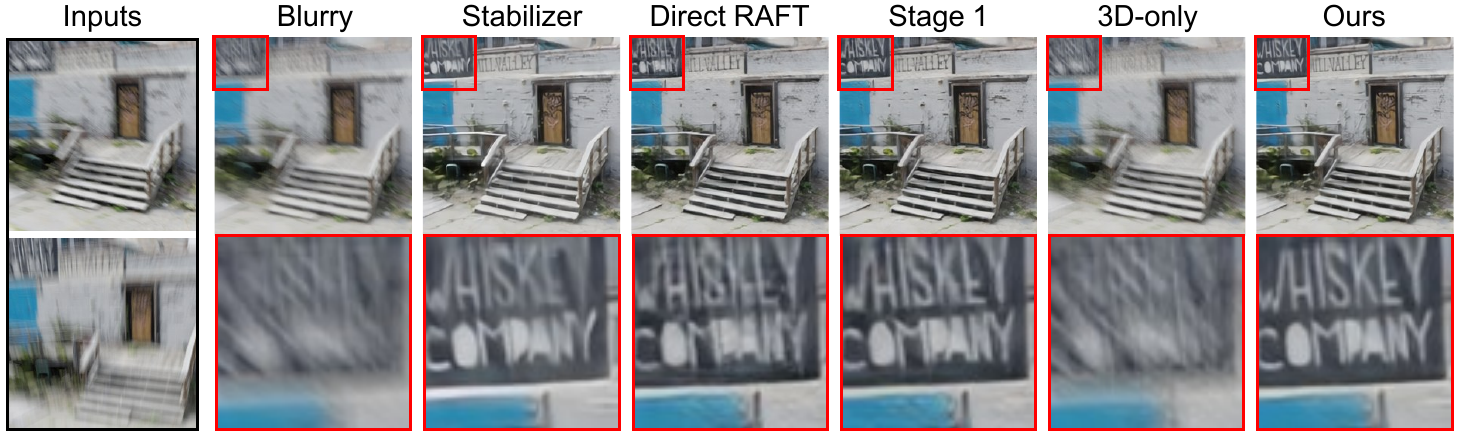}
    \caption{
Qualitative novel-view synthesis results for component diagnostics on a synthetic Deblur-NeRF scene. 
The Inputs column shows the two blurry views; for each reconstruction setting, the lower row enlarges the red-boxed region.
    }
    \label{fig:supp_component_diagnostics}
\end{figure}

\Cref{tab:supp_component_diagnostics,fig:supp_component_diagnostics} provide complementary quantitative and qualitative diagnostics beyond the ablation in the main paper.

\paragraph{Effect of alignment preconditioning.}
We estimate RAFT flows and construct OCGM guidance directly from the original blurry views while reusing the same Stage~1 and Stage~2 checkpoints as the full model.
This inference-time diagnostic removes the stabilizer only from the flow and guidance construction path, without retraining either restoration network.
Compared with the full cascade, PSNR and SSIM decrease by 0.33 dB and 0.020, respectively.
These results support the role of the stabilizer in producing alignment-friendly observations for reliable correspondence guidance, rather than serving as the final restoration module.

\paragraph{Effect of global 3D guidance without Stage~1.}
We additionally evaluate an inference-time 3D-only diagnostic that bypasses the stabilizer, RAFT, OCGM, and the entire Stage~1 restoration process.
The original blurry input pair is first passed to the frozen NoPoSplat backbone to produce input-view re-render guidance.
The blurry inputs and corresponding re-renders are then processed by the existing Stage~2 checkpoint, and the resulting restorations are used for final reconstruction with the frozen NoPoSplat backbone.

This configuration obtains 20.26 dB PSNR and 0.627 SSIM, substantially underperforming both Stage~1 and the full cascade.
The result indicates that global 3D re-render guidance derived directly from blurry inputs cannot replace the locally reliable observations established by Stage~1.
Because the Stage~2 checkpoint is reused without retraining, this experiment should be interpreted as an inference-time diagnostic
rather than an optimized 3D-only architecture.

\paragraph{Qualitative interpretation.}
\Cref{fig:supp_component_diagnostics} further illustrates how each diagnostic configuration affects the final novel-view reconstruction.
Independent stabilization improves each view separately but does not explicitly enforce cross-view consistency, leaving residual
discrepancies that are difficult for the frozen pose-free 3DGS backbone to aggregate coherently.
Direct RAFT estimates correspondences directly from the original blurry inputs, producing less reliable OCGM guidance.
This degrades the Stage~1 restorations and consequently the 3D guidance available to Stage~2.

Stage~1 combines stabilization and OCGM to establish more reliable local correspondences and therefore yields a more coherent reconstruction.
However, it lacks the global 3D feedback provided by Stage~2, leaving residual inconsistencies that cannot be resolved through local 2D warping alone.
Conversely, 3D-only constructs global guidance directly from the blurry inputs, before locally reliable observations have been
established.
The full cascade addresses both limitations by first improving local cross-view reliability and then refining the restored views using globally aggregated 3D guidance.

\begin{table}[t]
\caption{
Extended component diagnostics on real-world Deblur-NeRF scenes.
The first four rows reproduce the progressive ablation from the main
paper; the last two are inference-time diagnostics using the trained
checkpoints without retraining.
}
\label{tab:supp_component_diagnostics}
\centering
\small
\setlength{\tabcolsep}{4pt}
\renewcommand{\arraystretch}{1.05}

\begin{tabularx}{\linewidth}{
@{}
l
>{\raggedright\arraybackslash}X
cc
@{}
}
\toprule
Setting
& Configuration
& PSNR $\uparrow$
& SSIM $\uparrow$ \\
\midrule

\multicolumn{4}{@{}l}{\textit{Progressive cascade}} \\

Blurry inputs
& No restoration or cross-view guidance
& 20.07
& 0.612 \\

Stabilizer
& Independent per-view stabilization
& 20.30
& 0.655 \\

Stage~1
& Stabilizer and OCGM-based local 2D guidance
& 20.88
& 0.684 \\

Full cascade
& Complete CasDeblurGS pipeline
& \textbf{21.13}
& \textbf{0.700} \\

\midrule
\multicolumn{4}{@{}l}{\textit{Additional diagnostics}} \\

Direct RAFT
& RAFT and OCGM applied directly to the original blurry views
& 20.80
& 0.680 \\

3D-only
& Stage~1 bypassed; Stage~2 uses blurry inputs and 3D re-renders
& 20.26
& 0.627 \\

\bottomrule
\end{tabularx}
\end{table}

\section{Sparse Camera Reprojection Analysis}
\label{sec:supp_reprojection}

We further describe the sparse reprojection analysis reported in the main paper.
We compare the intermediate restorations produced by Stage~1 with the final restorations produced by Stage~2 using the same fixed real-world two-view pairs.

For each restored image pair, we extract OpenCV SIFT features with at most 4,000 keypoints and perform brute-force matching using the $\ell_2$ distance.
We apply a Lowe ratio threshold of 0.75 and retain only bidirectionally mutual matches.
The resulting correspondences are triangulated using \texttt{cv2.triangulatePoints} and the benchmark camera projection matrices.

A triangulated point is considered valid if its homogeneous coordinates are finite, its homogeneous denominator is nonzero, and it has positive depth in both camera frames.
For each valid point, we compute the symmetric reprojection error as the average pixel $\ell_2$ error after reprojecting the point into the two input views.

For each evaluation pair, we measure the number of valid triangulated points, the median symmetric reprojection error, the number of points with reprojection error below one pixel, and the corresponding inlier ratio.
The values reported in the main paper are obtained by averaging these per-pair measurements over the 60 real-world evaluation pairs.
The reported inlier ratio is therefore the mean of the per-pair ratios, rather than the mean inlier count divided by the mean number of valid points.
Benchmark camera poses are used only for this post-hoc evaluation and are never provided as inputs to CasDeblurGS.

\section{Runtime Analysis}
\label{sec:supp_runtime}

Although CasDeblurGS employs multiple frozen modules, all inference stages are feed-forward and require no per-scene optimization.
\Cref{tab:supp_runtime} reports average runtime, PSNR, and SSIM across both the real-world and synthetic Deblur-NeRF subsets.
CasDeblurGS improves PSNR by 4.48 dB over GAURA, the fastest baseline, while running 7.5\% faster than Difix3D+ and 31.8\% faster than CoherentGS, the closest competing methods in reconstruction quality.
These results demonstrate a favorable trade-off between reconstruction quality and computational cost.

\begin{table}[h]
\caption{
Average runtime and reconstruction quality across both the real-world and synthetic Deblur-NeRF subsets.
DAVANet and Difix3D+ are combined with the same frozen NoPoSplat backbone used in CasDeblurGS.
Best and second-best results are highlighted in \textbf{bold} and \underline{underlined}, respectively.
}
\label{tab:supp_runtime}
\centering
\small
\setlength{\tabcolsep}{4pt}
\renewcommand{\arraystretch}{1.05}
\begin{tabular*}{\columnwidth}{
@{\extracolsep{\fill}}
lccc
@{}
}
\toprule
Method
& Runtime (s) $\downarrow$
& PSNR $\uparrow$
& SSIM $\uparrow$ \\
\midrule
SE-GS~\cite{zhao2025self}
& \underline{28.48}
& 17.33
& 0.479 \\

GAURA~\cite{gupta2024gaura}
& \textbf{25.24}
& 17.79
& 0.516 \\

CoherentGS~\cite{xu2025breaking}
& 96.14
& 20.51
& \underline{0.674} \\

DAVANet~\cite{zhou2019davanet} + NoPoSplat
& 59.37
& 19.48
& 0.599 \\

Difix3D+~\cite{wu2025difix3d+} + NoPoSplat
& 70.85
& \underline{20.62}
& 0.642 \\

\textbf{CasDeblurGS (Ours)}
& 65.52
& \textbf{22.27}
& \textbf{0.748} \\
\bottomrule
\end{tabular*}
\end{table}

\section{Additional Experimental Results}
\label{sec:additional_results}

In this section, we provide per-scene quantitative evaluations and additional qualitative comparisons on Deblur-NeRF to further support the findings in the main paper.

\paragraph{Per-scene quantitative results.}
\Cref{tab:additional_results_synthetic,tab:additional_results_real} present per-scene quantitative comparisons on the synthetic and real-world Deblur-NeRF datasets, respectively. 
Consistent with the average gains reported in the main paper, our method achieves the best average performance and remains competitive across individual scenes.
These results show that the proposed cascaded 2D-to-3D guidance performs consistently across diverse scene structures and blur patterns.

\begin{figure}[tb]
    \centering
    \includegraphics[width=\linewidth]{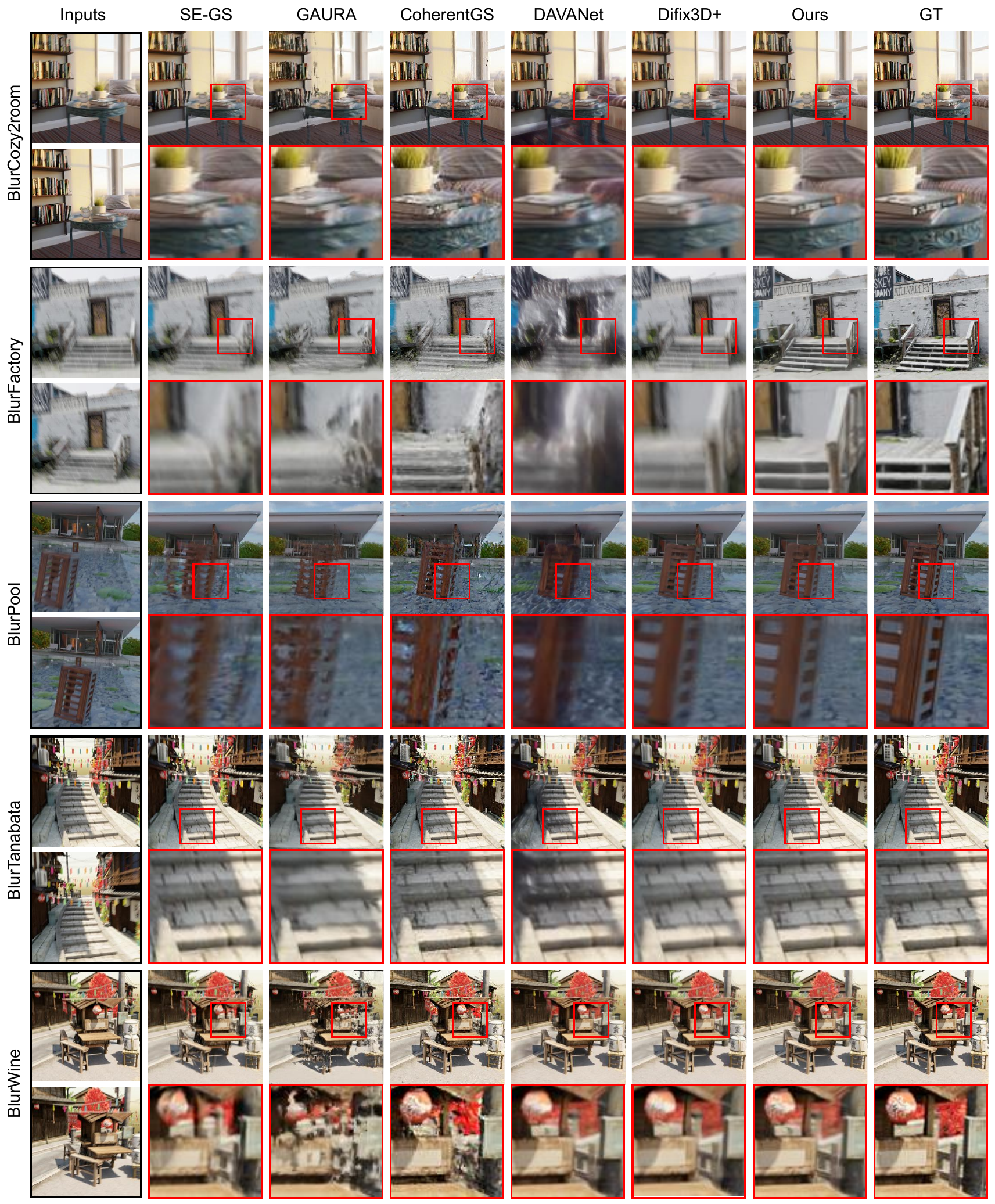}
    \caption{
        Additional qualitative results of novel view synthesis on synthetic scenes of the Deblur-NeRF dataset.
        The Inputs column shows the two blurry views; for each method, the lower row enlarges the red-boxed region.
    }
    \label{fig:additional_results_synthetic}
\end{figure}

\begin{figure}[tb]
    \centering
    \includegraphics[width=\linewidth]{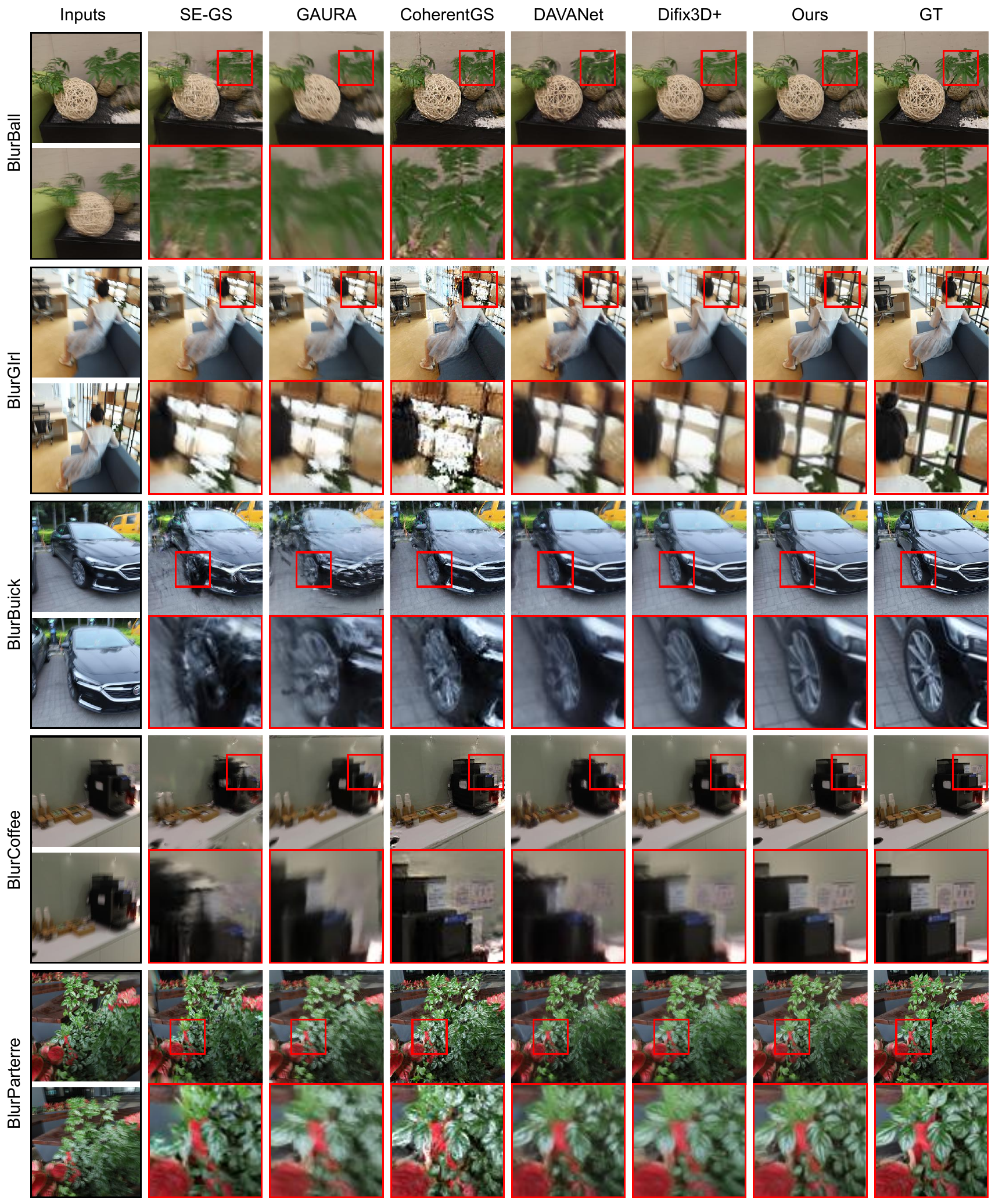}
    \caption{
        Additional qualitative results of novel view synthesis on real-world scenes of the Deblur-NeRF dataset.
        The Inputs column shows the two blurry views; for each method, the lower row enlarges the red-boxed region.
    }
    \label{fig:additional_results_real}
\end{figure}

\paragraph{Additional qualitative results.}
\Cref{fig:additional_results_synthetic,fig:additional_results_real} provide additional qualitative comparisons that further support the quantitative results. 
Under the challenging two-view setting with severe blur, existing methods often suffer from structural collapse or ghosting artifacts. 
For example, in the \textit{BlurFactory} scene, our framework reconstructs the staircase with more coherent geometry and finer texture details, whereas prior 3DGS-based methods struggle to preserve its topology. 
In the \textit{BlurCoffee} scene, CoherentGS exhibits noticeable color artifacts, while Difix3D+ produces overly blurred results; by contrast, our method recovers clearer text on the notice.
Overall, by combining local 2D correspondence guidance with global 3D re-render guidance, our approach improves multi-view consistency and yields higher-quality novel-view synthesis, especially in geometrically challenging regions.

\section{Supplementary Video}
\label{sec:supp_video}

The supplementary video provides an animated overview of CasDeblurGS, illustrating how local 2D correspondence guidance is progressively complemented by global 3D re-render guidance.
It also presents novel-view synthesis results from our method on nine scenes spanning the real-world and synthetic Deblur-NeRF subsets, enabling temporal inspection of rendering sharpness and structural consistency beyond the still-image results in the paper.


\begin{table}[b]
\caption{Quantitative results of novel view synthesis on synthetic scenes of the Deblur-NeRF dataset, reported for each individual scene. The best and second-best results are highlighted in \textbf{bold} and \underline{underlined}, respectively.}
\label{tab:additional_results_synthetic}
\centering
\resizebox{\linewidth}{!}{%
\begin{tabular}{llccc @{\qquad} llccc}
\toprule
Method & Scene & PSNR $\uparrow$ & SSIM $\uparrow$ & LPIPS $\downarrow$ & Method & Scene & PSNR $\uparrow$ & SSIM $\uparrow$ & LPIPS $\downarrow$ \\
\midrule
\multirow{6}{*}{SE-GS} 
 & BlurCozy2room & 20.65 & 0.686 & 0.209 & \multirow{6}{*}{DAVANet} 
 & BlurCozy2room & 21.83 & 0.770 & 0.185 \\
 & BlurFactory & 17.36 & 0.420 & 0.487 & 
 & BlurFactory & 16.27 & 0.385 & 0.488 \\
 & BlurPool & 22.87 & 0.660 & 0.265 & 
 & BlurPool & 21.53 & 0.607 & 0.270 \\
 & BlurTanabata & 15.72 & 0.497 & 0.377 & 
 & BlurTanabata & 18.93 & 0.673 & 0.268 \\
 & BlurWine & 17.19 & 0.536 & 0.353 & 
 & BlurWine & 19.50 & 0.691 & 0.247 \\
\cmidrule(lr){2-5} \cmidrule(lr){7-10}
 & Average & 18.76 & 0.560 & 0.338 & 
 & Average & 19.61 & 0.625 & 0.292 \\
\midrule
\multirow{6}{*}{GAURA} 
 & BlurCozy2room & 17.85 & 0.604 & 0.345 & \multirow{6}{*}{Difix3D+} 
 & BlurCozy2room & 23.60 & 0.801 & 0.138 \\
 & BlurFactory & 16.96 & 0.409 & 0.504 & 
 & BlurFactory & 18.12 & 0.479 & 0.457 \\
 & BlurPool & 23.04 & 0.625 & 0.331 & 
 & BlurPool & 26.33 & 0.742 & 0.168 \\
 & BlurTanabata & 16.08 & 0.526 & 0.431 & 
 & BlurTanabata & 19.10 & 0.682 & 0.277 \\
 & BlurWine & 15.63 & 0.458 & 0.459 & 
 & BlurWine & 19.35 & 0.663 & 0.273 \\
\cmidrule(lr){2-5} \cmidrule(lr){7-10}
 & Average & 17.91 & 0.524 & 0.414 & 
 & Average & \underline{21.30} & 0.673 & 0.262 \\
\midrule
\multirow{6}{*}{CoherentGS} 
 & BlurCozy2room & 24.91 & 0.828 & 0.160 & \multirow{6}{*}{\textbf{Ours}} 
 & BlurCozy2room & 25.46 & 0.866 & 0.097 \\
 & BlurFactory & 18.55 & 0.530 & 0.385 & 
 & BlurFactory & 23.73 & 0.818 & 0.203 \\
 & BlurPool & 24.33 & 0.705 & 0.227 & 
 & BlurPool & 27.14 & 0.775 & 0.142 \\
 & BlurTanabata & 18.56 & 0.674 & 0.276 & 
 & BlurTanabata & 19.99 & 0.748 & 0.202 \\
 & BlurWine & 19.21 & 0.696 & 0.247 & 
 & BlurWine & 20.75 & 0.771 & 0.182 \\
\cmidrule(lr){2-5} \cmidrule(lr){7-10}
 & Average & 21.11 & \underline{0.687} & \underline{0.259} & 
 & Average & \textbf{23.41} & \textbf{0.796} & \textbf{0.165} \\
\bottomrule
\end{tabular}%
}
\end{table}

\begin{table}[tb]
\caption{Quantitative results of novel view synthesis on real-world scenes of the Deblur-NeRF dataset, reported for each individual scene. The best and second-best results are highlighted in \textbf{bold} and \underline{underlined}, respectively.}
\label{tab:additional_results_real}
\centering
\resizebox{\linewidth}{!}{
\begin{tabular}{llccc @{\qquad} llccc} 
\toprule
Method & Scene & PSNR $\uparrow$ & SSIM $\uparrow$ & LPIPS $\downarrow$ & Method & Scene & PSNR $\uparrow$ & SSIM $\uparrow$ & LPIPS $\downarrow$ \\
\midrule
\multirow{11}{*}{SE-GS} 
 & BlurBall & 18.80 & 0.518 & 0.404 & \multirow{11}{*}{DAVANet} 
 & BlurBall & 20.74 & 0.598 & 0.342 \\
 & BlurBasket & 14.91 & 0.358 & 0.479 & 
 & BlurBasket & 19.54 & 0.578 & 0.378 \\
 & BlurBuick & 13.93 & 0.376 & 0.467 & 
 & BlurBuick & 18.46 & 0.602 & 0.340 \\
 & BlurCoffee & 18.74 & 0.677 & 0.351 & 
 & BlurCoffee & 23.11 & 0.779 & 0.280 \\
 & BlurDecoration & 14.10 & 0.281 & 0.495 & 
 & BlurDecoration & 16.88 & 0.457 & 0.411 \\
 & BlurGirl & 15.32 & 0.530 & 0.359 & 
 & BlurGirl & 19.35 & 0.713 & 0.291 \\
 & BlurHeron & 15.97 & 0.334 & 0.413 & 
 & BlurHeron & 18.85 & 0.487 & 0.390 \\
 & BlurParterre & 16.58 & 0.313 & 0.454 & 
 & BlurParterre & 19.30 & 0.468 & 0.406 \\
 & BlurPuppet & 14.18 & 0.284 & 0.505 & 
 & BlurPuppet & 18.76 & 0.524 & 0.343 \\
 & BlurStair & 16.45 & 0.304 & 0.499 & 
 & BlurStair & 18.42 & 0.527 & 0.415 \\
\cmidrule(lr){2-5} \cmidrule(lr){7-10}
 & Average & 15.90 & 0.398 & 0.443 & 
 & Average & 19.34 & 0.573 & 0.360 \\
\midrule
\multirow{11}{*}{GAURA} 
 & BlurBall & 20.11 & 0.569 & 0.433 & \multirow{11}{*}{Difix3D+} 
 & BlurBall & 21.35 & 0.622 & 0.309 \\
 & BlurBasket & 13.47 & 0.406 & 0.499 & 
 & BlurBasket & 19.98 & 0.622 & 0.326 \\
 & BlurBuick & 14.30 & 0.383 & 0.506 & 
 & BlurBuick & 18.45 & 0.621 & 0.293 \\
 & BlurCoffee & 22.53 & 0.800 & 0.310 & 
 & BlurCoffee & 23.74 & 0.815 & 0.203 \\
 & BlurDecoration & 16.21 & 0.433 & 0.490 & 
 & BlurDecoration & 17.69 & 0.497 & 0.369 \\
 & BlurGirl & 17.23 & 0.632 & 0.381 & 
 & BlurGirl & 19.52 & 0.768 & 0.227 \\
 & BlurHeron & 17.57 & 0.446 & 0.451 & 
 & BlurHeron & 18.47 & 0.493 & 0.373 \\
 & BlurParterre & 18.86 & 0.451 & 0.461 & 
 & BlurParterre & 19.73 & 0.508 & 0.366 \\
 & BlurPuppet & 17.35 & 0.467 & 0.454 & 
 & BlurPuppet & 19.25 & 0.567 & 0.299 \\
 & BlurStair & 19.10 & 0.492 & 0.456 & 
 & BlurStair & 21.25 & 0.595 & 0.346 \\
\cmidrule(lr){2-5} \cmidrule(lr){7-10}
 & Average & 17.67 & 0.508 & 0.444 & 
 & Average & \underline{19.94} & 0.611 & 0.311 \\
\midrule
\multirow{11}{*}{CoherentGS} 
 & BlurBall & 21.75 & 0.672 & 0.303 & \multirow{11}{*}{\textbf{Ours}} 
 & BlurBall & 22.87 & 0.730 & 0.228 \\
 & BlurBasket & 20.04 & 0.668 & 0.281 & 
 & BlurBasket & 21.00 & 0.703 & 0.234 \\
 & BlurBuick & 16.61 & 0.546 & 0.357 & 
 & BlurBuick & 20.17 & 0.709 & 0.217 \\
 & BlurCoffee & 24.78 & 0.826 & 0.247 & 
 & BlurCoffee & 25.21 & 0.859 & 0.156 \\
 & BlurDecoration & 16.69 & 0.517 & 0.379 & 
 & BlurDecoration & 17.97 & 0.545 & 0.309 \\
 & BlurGirl & 19.04 & 0.734 & 0.256 & 
 & BlurGirl & 21.56 & 0.843 & 0.158 \\
 & BlurHeron & 18.88 & 0.599 & 0.321 & 
 & BlurHeron & 20.42 & 0.651 & 0.260 \\
 & BlurParterre & 19.72 & 0.618 & 0.306 & 
 & BlurParterre & 20.41 & 0.586 & 0.281 \\
 & BlurPuppet & 19.64 & 0.666 & 0.267 & 
 & BlurPuppet & 19.87 & 0.633 & 0.232 \\
 & BlurStair & 21.88 & 0.755 & 0.204 & 
 & BlurStair & 21.82 & 0.736 & 0.209 \\
\cmidrule(lr){2-5} \cmidrule(lr){7-10}
 & Average & 19.90 & \underline{0.660} & \underline{0.292} & 
 & Average & \textbf{21.13} & \textbf{0.700} & \textbf{0.228} \\
\bottomrule
\end{tabular}%
}
\end{table}

\clearpage

\bibliographystyle{splncs04}
\bibliography{main}